\documentclass[a4paper,fleqn]{cas-sc}

\usepackage[authoryear,longnamesfirst]{natbib}

\usepackage{color}

\def\tsc#1{\csdef{#1}{\textsc{\lowercase{#1}}\xspace}}
\tsc{WGM}
\tsc{QE}
\tsc{EP}
\tsc{PMS}
\tsc{BEC}
\tsc{DE}

\usepackage{array} 
\usepackage{graphicx} 
\usepackage{amssymb} 
\usepackage{pdflscape} 
\usepackage{rotating} 

\begin{document}
\let\WriteBookmarks\relax
\def\floatpagepagefraction{1}
\def\textpagefraction{.001}

\shorttitle{Review of Cross-Target Stance Detection}

\shortauthors{Jamadi Khiabani \& Zubiaga}

\title [mode = title]{Cross-Target Stance Detection: A Survey of Techniques, Datasets, and Challenges}                      



%
\author[1]{Parisa Jamadi Khiabani}[orcid=0000-0002-0433-5196]
\cormark[1]

\fnmark[1]

\ead{p.jamadikhiabani@qmul.ac.uk}


\credit{Investigation, Conceptualization, Data Curation, Formal Analysis, Writing - Original Draft}

\affiliation[1]{organization={Queen Mary University of London},
    addressline={Mile End Road}, 
    city={London},
    postcode={E1 4NS}, 
    country={United Kingdom}}

\author[1]{Arkaitz Zubiaga}[orcid=0000-0003-4583-3623]


\credit{Supervision, Writing - Review \& Editing}




\cortext[cor1]{Corresponding author}



\begin{abstract}
Stance detection is the task of determining the viewpoint expressed in a text towards a given target. A specific direction within the task focuses on cross-target stance detection, where a model trained on samples pertaining to certain targets is then applied to a new, unseen target. With the increasing need to analyze and mining viewpoints and opinions online, the task has recently seen a significant surge in interest. This review paper examines the advancements in cross-target stance detection over the last decade, highlighting the evolution from basic statistical methods to contemporary neural and LLM-based models. These advancements have led to notable improvements in accuracy and adaptability. Innovative approaches include the use of topic-grouped attention and adversarial learning for zero-shot detection, as well as fine-tuning techniques that enhance model robustness. Additionally, prompt-tuning methods and the integration of external knowledge have further refined model performance. A comprehensive overview of the datasets used for evaluating these models is also provided, offering valuable insights into the progress and challenges in the field. We conclude by highlighting emerging directions of research and by suggesting avenues for future work in the task.
\end{abstract}




\maketitle

\section{Introduction}
In today's information-centric world, information is increasingly available through the Internet and social media platforms, much of which contains opinionated content reflecting people's views. This presence continues to expand alongside the rising popularity of social media platforms, which, as per recent statistics, are utilized by over 80\% of the UK's population in 2024\footnote{\url{https://datareportal.com/reports/digital-2024-united-kingdom}}, which is also garnering significant public attention \citep{tian2020early}. However, due to the large volume of posts, monitoring opinions expressed on social media platforms remains a cumbersome and often infeasible task if done manually, necessitating automated assistance \citep{liu2022sentiment,zubiaga2019mining}. Consequently, there is a pressing need for innovative and improved methods to automatically classify and process these texts, discerning the stance conveyed within, with the ultimate goal of mining public opinion. Indeed, social media provide access to unprecedented volumes of information from a diversity of users to form an estimate of the public opinion on a particular matter \citep{dong2021review}.

The rapid growth of social media platforms has brought forth new challenges in the field of information processing \citep{jamadi2020improved}. One prominent research area addressing opinions within social networks is Stance Detection (SD), which has the aim of forecasting the stance expressed in a text toward a specific entity \citep{biber1988adverbial}. Monitoring individual opinions or broader trends within communities and populations can yield valuable insights, particularly in domains such as politics, where it enables swift comprehension of public support towards certain topics or prediction of voting behavior.

In the stance detection problem, the input typically consists of a pair comprising a text and a target, and the output is a category chosen from the following 3-way set: {Favor, Against, None} \citep{alturayeif2023systematic}, or at times only two categories: {Favor, Against}. For example, for a text saying that ``I believe that climate change is a hoax,'' and where the target is the ``societal impact of climate change'', the stance would be ``None'' as the text indicates that the author refuses the importance of climate change. On occasions, some researchers expand this 3-way set to also include the category ``Neutral," indicating that the author shows a neutral stance toward the target \citep{grimminger2021hate}. However, it is also argued that a truly neutral stance is rare, as people generally lean either in favor of or against a proposition \citep{jaffe2009stance}. Moreover, the consensus in the literature suggests that if a text's stance toward a target is neither favorable nor opposing, the appropriate category should be ``None" rather than ``Neutral," since the text does not provide any clear stance information. Consequently, the ``None" category is typically used for all cases that do not fall into the Favor or Against categories.

\begin{figure}[htb]
	\centering
		\includegraphics[width=0.9\textwidth]{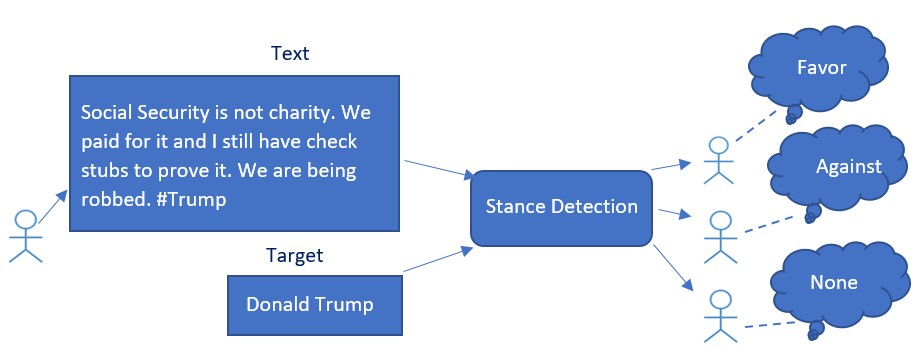}
	\caption{Stance detection scheme and its labels.}
	\label{figs:stance-detection.jpg}
\end{figure}

It should be noted that the target is anything for which a stance can be expressed, such as an entity, concept, event, idea, opinion, claim, or topic —-whether explicitly stated or implied within the text \citep{mohammad2016dataset,sobhani2017stance}. For example, in the text ``I consider myself pro-life'', the target topic ``abortion'' is implicit. The general scheme for 3-way stance detection is provided in Figure \ref{figs:stance-detection.jpg}. While the stance detection task has been popular for a relatively long period, one of the emerging foci over the last decade has been on cross-target stance detection (i.e. determining the stance for targets not seen during training), which is the focus of discussion of this survey paper.

\subsection{Scope of the Survey}

This survey covers cross-target and zero-shot stance detection, focusing on publicly available datasets and methods that advance these research areas. The included datasets were selected based on relevance and impact, particularly for their suitability in cross-target scenarios and zero-shot configurations. Our selection process involved searching Google Scholar with keywords like ``cross-target stance detection," ``zero-shot stance detection," and ``stance detection datasets." We prioritized datasets that are frequently cited, publicly accessible, and cover diverse domains such as politics, health, and finance.

Datasets like SemEval-2016 Task 6, VAST, and P-Stance were chosen for their specific design to support cross-target or zero-shot stance detection, making them ideal for evaluating model generalization across unseen targets. Datasets like RumourEval and COVID-19, with their varied topics and annotations, offer robust benchmarks for models to handle real-world complexities. This survey aims to provide a structured overview of these datasets and to highlight their roles in advancing stance detection research.

In addition to datasets, this survey explores the evolution of cross-target stance detection methods since 2016. Over this period, significant advancements have been made, with researchers employing various strategies to enhance model performance across diverse datasets and scenarios. The survey categorizes existing approaches into five major types: (1) Statistics-based methods, (2) Fine-tuning-based methods, (3) Prompt-tuning-based methods, (4) Knowledge-enhanced methods, and (5) Knowledge-enhanced Prompt-tuning methods. Each category represents a different approach to tackling the challenges of cross-target stance detection, reflecting the innovative trends and research focus areas that have emerged in the field. This comprehensive overview provides insight into the current landscape of cross-target stance detection, highlighting both foundational and cutting-edge methods.

\subsection{Related Surveys}

Several surveys have explored stance detection from various angles. Al-Dayel et al. offer a broad review across NLP, computational social science, and web science, covering foundational aspects and various methods, including network and contextual features. Their focus spans from traditional techniques to emerging trends in stance detection \citep{aldayel2021stance}.

Alturayeif et al. provide an overview of traditional stance detection methods and datasets up to 2022, emphasizing statistical models and foundational neural networks. They set the groundwork for understanding stance detection but do not delve deeply into the most recent advancements \citep{alturayeif2023systematic}.

Motyka et al. focus on target-phrase stance detection and zero-shot learning, addressing dataset limitations and comparing state-of-the-art methods. Their work highlights prompt-based learning and dataset inconsistencies \citep{motyka2024target}.

In contrast, our review paper is the first with a focus on cross-target stance detection, providing a detailed examination of advancements over the last decade and primarily from 2016 to 2024. We emphasize recent innovations such as zero-shot detection, reviewing fine-tuning, prompt-tuning, and knowledge-enhanced techniques. Our review categorizes methodologies into five distinct types and explores the evolution of cross-target stance detection more comprehensively, offering a nuanced and current perspective on the field.

\section{Background on Stance Detection}

\subsection{Problem Formulation of Stance Detection}

Stance Detection is a crucial task within Online Social Network (OSN) analysis, aiming to discern whether individuals express a favorable, opposing, or neutral stance towards specific targets, which could range from personal opinions to institutional policies or product preferences. Users articulate their stances across various platforms like online forums, Twitter, YouTube, and Instagram \citep{sobhani2016detecting}. This process not only facilitates individuals in expressing their viewpoints but also enables the aggregation of valuable insights, spanning from individual preferences to organizational and governmental perspectives \citep{darwish2020unsupervised}. Stance Detection thus emerges as a vital component within opinion mining, closely aligned with the analysis of user sentiments across social media platforms.

Stance detection is defined as the process of categorizing each post in a collection $P = \{p_1, p_2, \ldots, p_n\}$ into one of three stances : ``favor", ``against", or ``none" in a 3-class setting and into one of two stances: ``favor" or ``against" in a 2-class setting. Each post $p_i$ expresses a stance towards a specific target $t$. Stance datasets consist of posts expressing stances towards targets in a collection $T = \{t_1, t_2, \ldots, t_m\}$.

The in-target stance detection is the process of training and testing the model on the same target $t_i$. However, the cross-target stance detection task involves predicting the stance expressed in posts referring to target $t_i$, where the training data is composed of posts referring to other targets excluding $t_i$, hence requiring a transfer of knowledge from one set of targets to another.

\subsection{Stance Detection vs Sentiment Analysis}

\begin{figure}[htb]
	\centering
		\includegraphics[width=0.85\textwidth]{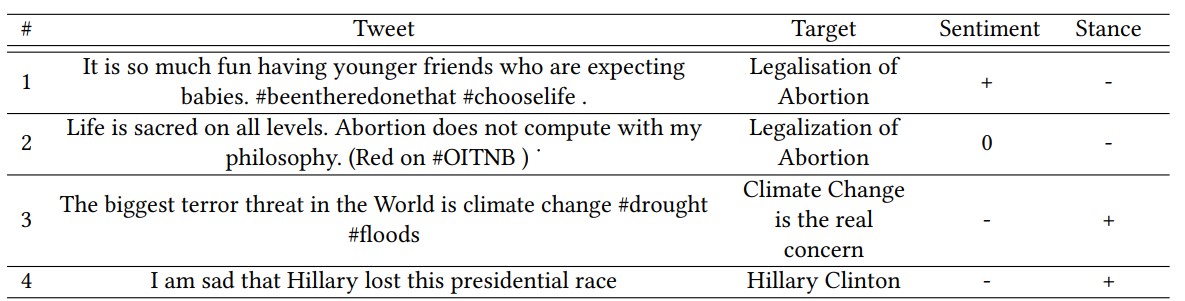}
	\caption{Some examples regarding the differences between sentiment and stance.}
	\label{figs:stance-sentiment.jpg}
\end{figure}

In the literature, stance detection is often deemed to be closely related to, and sometimes even conflated with, sentiment analysis. This is however often deemed to be a misconception of the concept of stance, and in fact research has shown that there are substantial differences between stance and sentiment \citep{aldayel2019assessing}. Indeed, sentiment analysis deals with identifying the explicit sentiment polarity expressed in a text, typically categorized as Positive, Negative, or Neutral. In contrast, stance detection seeks to determine the stance of a text towards a target, such as an event, entity, idea, claim, or topic \citep{alturayeif2023systematic}. For example, the text ``I'm glad that Donald Trump lost the election'' indicates a positive overall sentiment, whereas the stance towards Donald Trump is negative. The key distinctions between sentiment analysis and stance detection are: (1) sentiment analysis addresses the overall sentiment of the text often without a specific target, which is necessary in stance detection, and (2) the sentiment and stance towards a target within the same text might not match. For example, the text could have a positive overall sentiment while expressing a negative stance towards a particular target, or the reverse \citep{kuccuk2020stance}. Figure \ref{figs:stance-sentiment.jpg} shows examples with their sentiment and stance labels \citep{aldayel2021stance}.

Two subproblems of sentiment analysis are notably closer to stance detection than the broader sentiment analysis problem itself \citep{kuccuk2020stance}:

\textbf{Aspect-Oriented (or Aspect-Based, or Aspect-Level) Sentiment Analysis:} This subproblem delves into the sentiment polarities towards a target entity and its various aspects within a given text input \citep{pontiki2016semeval,schouten2015survey}. Typically, it is treated as a slot-filling task involving three slots: the target entity, the aspect of the entity, and the sentiment polarity towards that aspect. Commonly examined target entities in shared datasets for aspect-oriented sentiment analysis include electronic devices like laptops, restaurants, and hotels, while corresponding aspects might include price, design, and quality, among others.

\textbf{Target-Dependent (or Target-Based) Sentiment Analysis:} This subproblem focuses on determining the sentiment polarity towards a specific target entity within the text, given a text and target pair \citep{jiang2011target}. A similar variant is open-domain targeted sentiment analysis, where both a named entity and the sentiment toward that entity are explored in the input text \citep{mitchell2013open}. As outlined in \citet{ebrahimi2016joint}, key distinctions between stance detection and target-dependent sentiment analysis include: (1) the stance target may not be explicitly provided in the input text, (2) the stance target may not necessarily be the target of the sentiment in the text, and (3) while the stance target could be an event, targets in sentiment analysis are typically entities or aspects. These differences similarly extend to stance detection and open-domain targeted sentiment analysis.

\section{Stance datasets}
\label{datasets}

In this section, we discuss publicly available datasets which have been released in the scientific literature, particularly focusing on those datasets that are suited to or have been used for cross-target stance detection. As a summary, Table \ref{tab:datasets} presents a comparative analysis of the most widely used stance detection datasets, which we discuss in more detail next. 

\textbf{SemEval-2016 Task 6 (Sem2016T6):}
This dataset presented by Mohammad et al.  focused on detecting the stance expressed in tweets towards predefined targets. The challenge was divided into two subtasks: Task A, a supervised stance detection task with five targets—Atheism, Climate Change is a Real Concern, Feminist Movement, Hillary Clinton, and Legalization of Abortion—and Task B, a weakly supervised task centered on a single target, Donald Trump, for which no labeled training data was provided. In Task A, approximately 70\% of the data was used for training and 30\% for testing. This task saw 19 submissions, with the highest F-score reaching 67.82. Task B, dealing with the lack of training data, had 9 submissions and achieved a top F-score of 56.28, illustrating the challenge of stance detection without explicit training data \citep{mohammad2016semeval}.

The dataset comprised 4,870 annotated tweets categorized into `favor', `against', and `neither'. Tweets were collected using target-related hashtags and manually annotated by crowd workers. The evaluation employed macro-average F1-scores for the `favor' and `against' classes, treating the `neither' class as non-interest in this context. This task highlighted the complexities of stance detection, particularly when the target is not directly mentioned, and aimed to advance research by providing data and tools for further exploration in stance detection and related areas. The dataset is suited for both stance detection and cross-target stance detection, evaluating performance on the remaining target in the zero-shot configuration.

\textbf{Emergent}: The Emergent dataset is a valuable asset developed from a digital journalism initiative aimed at combating misinformation. It encompasses 300 rumor claims and 2,595 related news articles, with each claim and article meticulously annotated by journalists. The dataset categorizes the claims into three veracity labels: true, false, or unverified. Each news article is further summarized with a headline and labeled according to its stance toward the associated claim. The stance labels—``for," ``against," and ``observing"—indicate whether the article supports, refutes, or simply reports the claim without evaluation. This comprehensive approach provides a rich source for various natural language processing applications, especially those focused on fact-checking \citep{ferreira2016emergent}.

Emergent stands out due to its emphasis on stance classification at the headline level, capturing how articles position themselves relative to the claims they discuss. This dataset includes a wide array of claims on topics ranging from global news to technology, facilitating thorough evaluation of stance detection methods. The dataset features a well-distributed range of stance labels—47.7\% for, 15.2\% against, and 37.1\% observing—which enhances the reliability of stance classification models. By integrating data from real-world journalism, Emergent offers a more authentic challenge for NLP techniques. Innovations in feature extraction, including syntactic analysis and word alignment, have resulted in improved accuracy for stance detection, positioning Emergent as a crucial tool for advancing automated fact-checking and computational journalism research.

\textbf{Multi-target}: The Multi-Target stance dataset is an innovative resource developed to address the challenge of stance detection towards multiple related targets within a single document. This dataset comprises 4,455 tweets, each annotated to reflect stances toward two specific targets simultaneously. Focused on the 2016 US presidential election, it includes tweets related to four candidates: Donald Trump, Hillary Clinton, Ted Cruz, and Bernie Sanders. The dataset is partitioned into training (70\%), development (10\%), and test (20\%) sets, and also provides a larger collection of unlabeled data for additional exploration. This resource is designed to capture the nuanced relationships and dependencies between stances on related targets, which is often overlooked in traditional stance detection models that treat each target independently \citep{sobhani2017dataset}.

The dataset's key innovation lies in its approach to modeling stance dependencies using advanced neural network architectures. The authors propose a sequence-to-sequence model with an attention mechanism to jointly predict stances for pairs of targets. This model outperforms simpler methods such as window-based or cascading classifiers by addressing the interdependencies between stance labels. By incorporating attention-based encoder-decoder frameworks, the dataset facilitates more accurate stance prediction and better reflects real-world complexities where opinions on multiple related targets are interrelated. This public dataset aims to advance research in multi-target stance detection and offers a robust foundation for developing and testing models that can handle interconnected subjectivities in various contexts.

\textbf{VAST:} This dataset presented by Allaway et al. includes 18,515 comments from the New York Times ``Room for Debate" section, covering a wide array of topics such as politics (e.g., ``a Palestinian state"), education (e.g., ``charter schools"), and public health (e.g., ``childhood vaccination"). Each comment is annotated with one of three stance labels: pro, con, or neutral. This dataset is particularly valuable for zero-shot stance detection due to its broad range of topics and diverse expressions. Unlike existing datasets with a limited number of topics and expressions, VAST includes a variety of similar expressions (e.g., ``guns on campus" versus ``firearms on campus") to capture realistic linguistic variations. This variation helps address the challenge of evaluating zero-shot and few-shot stance detection by incorporating diverse topic representations \citep{allaway2020zero}.

To create VAST, comments were extracted from the Argument Reasoning Comprehension (ARC) Corpus, and topics were heuristically identified and verified using crowdsourcing. The dataset features multiple types of annotations: heuristically extracted topics, corrected topics provided by annotators, and additional topics listed by annotators. Neutral examples were also included to account for comments that do not convey a clear stance. The resulting dataset has a median of four unique topics per comment and is well-suited for developing models for zero-shot and few-shot stance detection, given its complexity and variety of topics.

\textbf{RumourEval:} This dataset, developed for the SemEval 2017 shared task, offers a comprehensive resource for studying misinformation on social media. This dataset is designed to address two primary challenges: stance classification and veracity prediction. It includes a substantial collection of tweets organized into tree-structured conversation threads. The dataset covers eight major events, such as the Charlie Hebdo shooting and the Ferguson unrest, and consists of 4,519 tweets for training and 1,080 tweets for testing. Each tweet in these threads is labeled with one of four stance categories—Support, Deny, Query, or Comment—in response to a rumour, and the dataset also includes veracity labels indicating whether the rumour is true, false, or unverified \citep{derczynski2017semeval}.

A notable innovation of RumourEval is its focus on the conversational context around rumours, which is captured through direct and nested replies. This approach allows for a nuanced analysis of how public discourse evolves around a rumour, providing valuable insights into public sentiment and engagement. Additionally, the dataset incorporates external context, such as Wikipedia articles, to aid in the veracity prediction task. The annotation process combines expert and crowdsourced labeling, ensuring high-quality data for both stance and veracity classification. This structured and contextually rich dataset sets a benchmark for future research in rumour detection and misinformation analysis.

\textbf{RumourEval 2019:} This dataset builds upon its predecessor from 2017 to advance the automated
verification of online rumours. It comprises a comprehensive collection of social media posts from both Twitter and Reddit, reflecting a broad spectrum of news events and public reactions. The dataset is structured around two main subtasks: stance detection and rumour verification. For Task A, which focuses on stance detection, the dataset comprises a total of 8,574 posts, including both training and test data. This dataset features tweets and Reddit posts, which are annotated with stances categorized as supporting, denying, querying, or commenting. Task B, dedicated to rumour verification, includes 446 posts, which are classified into three veracity categories: true, false, or unverified.  The dataset has been significantly expanded, now including 297 source tweets with associated discussions, and augmented by new tweets and Reddit posts, increasing the depth and diversity of the data \citep{gorrell2019semeval}.

Innovatively, RumourEval 2019 introduces a mix of new and existing sources, such as the inclusion of Reddit,
which provides longer and more complex discussions compared to Twitter. The dataset’s richness is further enhanced by the introduction of additional natural disaster-related tweets and comprehensive stance annotations collected through crowd sourcing, ensuring high-quality and nuanced data. This iteration also emphasizes the integration of stance information from Subtask A into the rumour verification process in Subtask B, allowing for a more sophisticated approach to assessing rumour accuracy. The expanded and diverse dataset reflects the growing importance of automating rumour detection and verification in the face of increasing misinformation on social media platforms.

\textbf{COVID-19}: This Stance dataset is a pivotal resource designed to facilitate stance detection in the context of the COVID-19 pandemic, capturing public sentiment on key issues such as health mandates. This dataset includes 6,133 manually annotated tweets centered on four controversial targets: ``Anthony S. Fauci, M.D.," ``Keeping Schools Closed," ``Stay at Home Orders," and ``Wearing a Face Mask." The tweets are categorized into three stance labels: in-favor, against, and neither, allowing for nuanced analysis of public opinion. The data collection process spanned from February to August 2020, utilizing a variety of pandemic-related keywords and hashtags to capture relevant tweets, ensuring the dataset's comprehensiveness and relevance \citep{glandt2021stance}.

One of the dataset's major innovations is its meticulous annotation process, which involved crowdsourcing through Amazon Mechanical Turk and rigorous agreement measures to ensure data quality. This dataset not only provides stance labels but also includes annotations for sentiment and the explicitness of the opinion, adding layers of complexity for model training. The COVID-19-Stance dataset serves as a challenging benchmark for stance detection, given the mixture of explicit and implicit opinions, along with varied sentiment expressions within the tweets. By establishing baseline results with state-of-the-art models and exploring methods like self-training and domain adaptation, the dataset significantly advances research in NLP, particularly in understanding public attitudes towards health measures during a global crisis.

\textbf{STANDER}: This dataset is a notable resource designed for stance detection (SD) and fine-grained evidence retrieval (ER) in news articles, encompassing 3,291 meticulously annotated pieces focused on four major U.S. healthcare mergers and acquisitions (M\&A). The mergers in question involve high-profile companies: UnitedHealth Group's acquisition of Change Healthcare (UH-CCH), Cigna's acquisition of Express Scripts (CIG-ES, CVS's purchase of Aetna (CVS-AET), and the merger between Centene and WellCare (CEN-WC). Each article within the dataset is labeled for stance concerning these M\&A activities—categorized as Support, Refute, Comment, or Unrelated. In addition to stance labeling, the dataset is unique in its provision of detailed evidence snippets, with explicit start and end indices, enhancing the granularity of evidence retrieval \citep{glandt2021stance}.

STANDER’s contributions to the field are significant due to its detailed annotation and alignment with existing Twitter datasets. By integrating authoritative news sources with user-generated content, STANDER addresses previous gaps in multi-genre stance detection. The dataset also features a balanced stance distribution naturally emerging from its content, reflecting the realistic complexity of public and journalistic opinion on M\&A topics. Furthermore, its diachronic analysis of annotator disagreements provides valuable insights into how evolving public sentiment and uncertainty influence stance classification, establishing STANDER as a challenging and informative benchmark for future research in stance detection and rumor verification.

\textbf{WT-WT}: The Will-They-Won’t-They (WT-WT) dataset is a substantial contribution to the field of stance detection, particularly within the context of financial rumor verification. Comprising 51,284 tweets, it stands as the largest publicly available stance detection dataset focused on user-generated content. The dataset is specifically curated to address the rumor verification task in mergers and acquisitions (M\&A), making it uniquely relevant for applications in the financial domain. Tweets within this dataset are annotated by domain experts, ensuring high-quality labeling across four distinct categories: support (indicating belief that the merger will happen), refute (expressing doubt about the merger), comment (neutral discussions about the merger), and unrelated (tweets not related to the merger). This meticulous annotation process and the size of the dataset make it an invaluable resource for training and evaluating stance detection models \citep{conforti2020will}.

One of the key innovations of the WT-WT dataset is its focus on the financial domain, where stance detection plays a crucial role in interpreting market sentiment and verifying the accuracy of circulating rumors. Additionally, the dataset’s multi-domain nature, encompassing various sectors such as healthcare and entertainment, allows for robust cross-domain analysis and model adaptation. The dataset also highlights a significant challenge for current models, with experimental results showing a noticeable gap between machine and human performance. This performance gap underlines the dataset's potential to drive future research in improving model accuracy, exploring cross-target and cross-domain generalization, and integrating linguistic and network-based features for more nuanced stance detection.

\textbf{P-Stance:} This dataset introduced by Li et al. is designed to address gaps in stance detection datasets by facilitating large-scale evaluations that require deeper semantic understanding. Comprising 21,574 English tweets, this dataset focuses on the political domain and includes three targets: Donald Trump, Joe Biden, and Bernie Sanders. Each tweet is annotated with a stance label toward one of these targets, with labels being favor or against. The primary motivation behind P-Stance is to provide a comprehensive benchmark for both in-target stance detection—where a classifier is trained and validated on the same target—and cross-target stance detection, where a model is adapted from one target to another \citep{li2021p}.

Moreover, P-Stance introduces a novel task: cross-topic stance detection, where a classifier trained on one topic must generalize to different topics within the same target. This setup aims to bridge the gap between previous datasets with limited scope and those requiring more nuanced semantic understanding. P-Stance enables robust evaluation through its extensive dataset size and the complexity of its annotations.

The dataset was collected using the Twitter streaming API with query hashtags to gather tweets related to the political figures in focus. A rigorous preprocessing step filtered out tweets with fewer than 10 or more than 128 words, removed duplicates and retweets, and ensured all tweets were in English. The final corpus includes around 2 million examples after pre-processing. Annotation was conducted via Amazon Mechanical Turk (AMT), with tweets labeled as ``Favor,” ``Against,” ``None,” or ``I don’t know.” Quality assurance measures included filtering annotators based on performance metrics and re-annotating data where necessary. The dataset's challenging nature arises from the more implicit references to targets and longer tweet lengths compared to previous datasets. This makes P-Stance a demanding benchmark for stance detection tasks.

\textbf{CoVaxNet}: This dataset is a pioneering multi-source, multi-modal, and multi-feature data repository designed to address COVID-19 vaccine hesitancy by integrating both online and offline information sources. The dataset encompasses a diverse array of data types including social media posts, fact-checking reports, COVID-19 statistics, U.S. Census data, government responses, and local news reports. Specifically, CoVaxNet includes over 1.8 million tweets categorized by pro-vaccine and anti-vaccine sentiments, 4,263 fact-checking reports, and 813 low-credibility sources, complemented by comprehensive offline datasets such as daily COVID-19 statistics, detailed census data, government policy records, and extensive local news coverage. This dataset represents a significant advancement in capturing the multifaceted nature of vaccine-related discourse and behaviors, providing a robust framework for analyzing the intersection of online sentiments and offline realities \citep{jiang2023covaxnet}.

Innovatively, CoVaxNet connects online and offline data through a geolocation-based approach, enabling researchers to study the impact of online discussions on offline vaccine uptake and vice versa. This integration facilitates a nuanced understanding of vaccine hesitancy by linking social media discourse with real-world COVID-19 statistics and demographic information. The dataset’s design not only supports a wide range of research applications, including misinformation detection and the exploration of structural inequalities, but also provides valuable insights into how various factors influence vaccine acceptance and policy responses. CoVaxNet's comprehensive nature and novel approach to data integration position it as a critical resource for advancing research and informing public health strategies.

\textbf{ISD}: The ISD dataset is designed for implicit stance detection, focusing on tweets where user stance is implied, typically through the use of hashtags rather than direct sentiment words. It was collected during the 2020 U.S. presidential election and includes about 762,255 tweets, with 6,027 labeled for either Donald Trump (DT) or Joe Biden (JB). Unlike other stance detection datasets, ISD requires models to deduce stance from context or hashtags \citep{huang2023knowledge}.

The dataset underwent thorough preprocessing to remove irrelevant hashtags, duplicates, and retweets, ensuring cleaner data. Each tweet is annotated with a stance: ``Favor," ``Against," or ``None," and contains hashtags that convey implicit stance information. This makes ISD a complex dataset, as models must interpret underlying meanings in hashtags to accurately determine stance.

\textbf{ProCons}: The ProCons dataset, is a large-scale Chinese dataset tailored for zero-shot stance detection tasks. It includes 245 submissions on diverse topics, totaling 32,667 Chinese posts, each labeled with a stance of ``for," ``against," or ``neutral." This dataset is specifically designed to test the performance of models in zero-shot scenarios, where models are required to detect stances without prior training on the specific topics present in the dataset \citep{wang2023quantifying}.

\textbf{C-STANCE}: C-STANCE is the first large-scale Chinese dataset for zero-shot stance detection (ZSSD), consisting of 48,126 annotated text-target pairs from Sina Weibo. It introduces two challenging subtasks: target-based ZSSD, where models are tested on unseen targets, and domain-based ZSSD, which evaluates models on targets from entirely new domains. The dataset includes a diverse range of targets, both noun-phrases and claims, allowing for multiple targets per text. C-STANCE is over twice the size of the English ZSSD VAST dataset and establishes a new benchmark for stance detection in Chinese. With baseline models achieving only 78.5\% F1 macro, C-STANCE offers a demanding test for model generalization and supports future research in zero-shot learning \citep{zhao2023c}.

\textbf{MT-CSD (Multi-Turn Conversation Stance Detection):} The MT-CSD (Multi-Turn Conversation Stance Detection) dataset represents a significant advancement in the field of Conversational Stance Detection (CSD). This dataset is the largest of its kind, comprising 15,876 meticulously annotated instances, offering a rich resource for stance detection research. It encompasses multiple targets, including ``Tesla," ``SpaceX," ``Donald Trump," ``Joe Biden," and ``Bitcoin," collected from Reddit, one of the most extensive forums for online discourse. The dataset is particularly notable for its depth of conversation, with 75.99\% of the data consisting of comments that extend beyond four conversational turns. This depth contrasts sharply with previous datasets like CANT-CSD, where only 6.3\% of the data extends beyond three turns \citep{niu2024challenge}.

MT-CSD introduces unique challenges for stance detection due to its emphasis on multi-turn conversations. It features implicit target references embedded within local sub-discussions, requiring sophisticated methods to capture both long- and short-range dependencies in the data. To tackle these challenges, the dataset is accompanied by a novel Global-Local Attention Network (GLAN) model, designed to address the complexities inherent in such deep conversational data. Additionally, MT-CSD's focus on implicit stance cues, coreference resolution, and contextual understanding pushes the boundaries of existing stance detection approaches, making it a critical resource for advancing research in this domain. The dataset's rigorous construction process, including a two-reviewer relevance check and strict preprocessing criteria, ensures the high quality and relevance of the data, further enhancing its utility for developing robust stance detection models.

The dataset for stance detection encompasses a diverse range of sources, targets, and types, enabling comprehensive research and model development in this field.

\begin{table}[ht]
    \centering
    \resizebox{\textwidth}{!}{%
        \begin{tabular}{>{\raggedright\arraybackslash}p{3cm} 
                        >{\centering\arraybackslash}p{1.2cm} 
                        >{\centering\arraybackslash}p{2.5cm} 
                        >{\centering\arraybackslash}p{2.5cm} 
                        >{\raggedright\arraybackslash}p{8.5cm}}
            \toprule
            Dataset & Language & Size & Domain & Targets \\
            \midrule
            SemEval Task 6.A (2016) & English & 4,870 & Social and political (Twitter) & ’Atheism’, ’Climate change': CC, ’Feminist movement’: FM, ’Hillary Clinton’: HC, ’Legalization of abortion’: LA \\
            \midrule
            Semeval Task 6.B (2016) & English & 707 labeled tweets and 78,000 unlabeled tweets & Political (Twitter) & Unlabeled tweets related to Trump \\ 
            \midrule
            Emergent (2016) & English & 300 claims and 2,595 headlines & News headlines & Claims extracted from rumour sites and Twitter \\
            \midrule
            Multi-target (2017) & English & 4,455 & Political (Twitter) & ’Clinton-Sanders’, ’Clinton-Trump’, ’Cruz-Trump’ \\
            \midrule
            VAST (2020) & English & 4,986 & Politics, education, and public health & A large range of topics from The New York Times ‘Room for Debate’ section, part of the Argument Reasoning Comprehension (ARC) Corpus \\
            \midrule
            RumourEval (2017) & English & 5,568 & Social (Twitter) & Rumorous tweets \\ 
            \midrule
            RumourEval (2019) & English & 8,574 & Twitter and Reddit & Various events \\
            \midrule
            COVID-19 (2021) & English & 6,133 & Twitter & 'Anthony S. Fauci, M.D', 'Keeping Schools Closed', 'Stay at Home Orders', and 'Wearing a Face Mask' \\
            \midrule
            STANDER (2020) & English & 3,291 & Social media and news & News articles focused on four major U.S. healthcare mergers and acquisitions (M\&A) \\
            \midrule
            WT-WT (2020) & English & 51,284 & Financial (Twitter) & Various targets about companies mergers and acquisitions: 'Cigna-Express Scripts', 'Aetna-Humana', 'CVS-Aetna', 'Anthem-Cigna', and 'Disney-Fox' \\
            \midrule
            P-Stance (2021) & English & 21,574 & Politics (Twitter) & ’Donald Trump’, ’Joe Biden’, and ’Bernie Sanders’ \\
            \midrule
            CoVaxNet (2023) & English & 21,574 & Politics (Twitter) & Social media posts, fact-checking reports, COVID-19 statistics, U.S. Census data, government responses, and local news reports \\
            \midrule
            ISD (2023) & English & 6,027 & Politics (Twitter) & 'Donald Trump' and 'Joe Biden' \\
            \midrule
            ProCons (2023) & Chinese & 21,574 & Weibo & Social media posts, fact-checking reports, COVID-19 statistics, U.S. Census data, government responses, and local news reports \\
            \midrule
            C-STANCE (2023) & Chinese & 48,126 & Politics (Twitter) & 40,204 responses and local news reports\\
            \midrule
            MT-CSD (2024) & English & 15,876 & Reddit & Multiple targets, including ``Tesla," ``SpaceX," ``Donald Trump," ``Joe Biden," and ``Bitcoin," U.S. Census data, government responses, and local news reports \\
            \bottomrule
        \end{tabular}
    }
    \caption{Stance Detection Datasets}
    \label{tab:datasets}
\end{table}

These datasets collectively represent a broad spectrum of stance detection challenges, from specific topics and events to more general claim-based stances, sourced from both social media and news articles. This variety enables researchers to develop and evaluate models across different contexts, enhancing the robustness and generalizability of stance detection systems.

\section{Delving into Stance Detection}

Before delving into the various forms of stance detection, it is essential to distinguish between the different levels at which stance detection is applied. In the literature, stance classification is applied at two levels:

\begin{itemize}
 \item \textbf{Statement level stance detection:} Here, the goal is to predict the stance expressed in a piece of text. This approach is common in NLP research, where features are extracted from text such as forum posts \citep{murakami2010support} or tweets \citep{mohammad2016semeval}.

 \item \textbf{User level stance detection:} Here, the objective is to predict a user's overall stance on a given topic, which can then be achieved by looking at multiple posts from a user, rather than a single post. This method can also incorporate various user attributes alongside the text in their posts \citep{aldayel2019your}.
\end{itemize}

Stance detection is crucial in analytical studies aimed at gauging public opinion on social media, especially on political and social issues. These issues are often contentious, leading people to express conflicting opinions on distinguishable points. Topics like abortion, climate change, and feminism are frequently used as focal points for stance detection on social media \citep{mohammad2016semeval}. Similarly, political matters, such as referenda and elections, are consistently popular domains for stance detection to explore public opinion \citep{fraisier2018stance}. Stance detection, also known as perspective \citep{klebanov2010vocabulary} and viewpoint \citep{trabelsi2018unsupervised,zhu2019hierarchical} detection, involves identifying perspectives by expressing stances on a contentious topic \citep{mohammad2017stance}.

Stance detection involves discerning an individual's or a post's perspective regarding a specific subject \citep{biber1988adverbial}. Typically, this perspective is categorized as either supportive or opposing towards the subject in question \citep{aldayel2021stance,jamadi2020improved,liu2022sentiment,biber1988adverbial,zubiaga2018longitudinal,mohammad2016semeval}. However, classifying stances from social media data poses a challenge \citep{antoun2020arabert, zhang2020enhancing}, primarily due to the varied and informal nature of such data. Detecting stances in Twitter posts presents distinct hurdles for researchers due to the platform's character limit, where tweets may often have limited context due to this brevity, and the informal nature of tweets. These posts, often abbreviated and lacking formal structure, frequently deviate from standard grammatical conventions \citep{siddiqua2019tweet}.

Considering the number of targets and whether the stance target appears in both the training and testing datasets, three subclasses of the stance detection problem can be identified: target-specific (in-target) stance detection \citep{mohammad2016semeval}, multi-target stance detection \citep{sobhani2017stance} and cross-target stance detection \citep{augenstein2016stance,xu2018cross}.  These three subtasks are defined as follows:

\begin{itemize}
 \item \textbf{In-target Stance Detection:} In studies focused on specific targets (in-target), the primary input involves either the text itself or user input, aiming to discern the stance towards predefined targets, like Donald Trump in the US election or the BREXIT referendum \citep{alturayeif2023systematic}. These types of techniques (In-target) are mainly concerned with inferring the stance for a set of predefined targets, building a separate stance classification model for each target \citep{aldayel2019your} \citep{siddiqua2018stance}. Current approaches have demonstrated encouraging performance in in-target stance detection, where the same targets appear in both the training and test datasets \citep{mohammad2016dataset}

 \item \textbf{Multi-target Stance Detection:} The core idea behind this type of stance detection is: ``when a person gives their stance for one target, it provides information on their stance toward other related targets" \citep{aldayel2021stance}. Therefore, a stance class towards multiple targets is considered in this kind of methods for a given piece of input. Hence, learning social media users’ tendency regarding two or more targets for a single topic is achievable \citep{sobhani2017dataset}. Take in-favor stance for Hillary Clinton as a target example, it shows an against stance toward Trump \citep{darwish2017trump}.
 \item \textbf{Cross-target Stance Detection:} As discussed earlier, most of the available stance detection approaches are concerned with in-target (target-specific) representations that models are trained and tested using data specific to the same target. But in some cases, the target may have few or no labeled data which is a limitation for in-target settings; for example, one may have a labelled dataset with stances towards Donald Trump, which may want to be exploited to classify the stances in texts towards Joe Biden, a target for which labelled data is not available. Recently, a growing body of work has been emerged to explore the concept cross-target representations on social media as a new frontier by ﬁne-tuning large pre-trained language models with a comparatively small portion of data, leading to distinguished performance enhancement for downstream NLP/NLU tasks. In other words, cross-target stance detection approaches are utilized to rectify the lack of labeled training data for new targets \citep{wei2019modeling}. These models will be built for a destination target using labeled data from a different target but related one, alleviating required annotation of new targets due to using labeled data associated with existing targets. Noteworthy is the fact that cross-target stance detection requires human knowledge about any new target and its relationship with the training targets \citep{allaway2020zero}. So target generalization is the main concern regarding cross-target settings. We believe that cross-target stance detection will be a fruitful line of future work.
\end{itemize}

Before delving into cross-target approaches in the next section, we will first define key terminologies and methodologies commonly used in the stance literature.

\textbf{Contrastive Learning:} Contrastive learning is a method focused on self-supervised representation learning. It seeks to position similar items (positive pairs) close together in the embedding space, while keeping dissimilar items (negative pairs) farther apart. This technique enhances the representation of data by maximizing the mutual information between different augmented versions of the same sample. Specifically, supervised contrastive learning refines this process by bringing examples of the same category closer and separating those from different categories, thereby creating a more effective semantic representation of the data \citep{chunling2023adversarial,zheng2022knowledge}.

\textbf{Domain Adaptation:} Domain Adaptation is a transfer learning technique aimed at bridging the gap between domains by reducing domain differences and improving model generalization. It is primarily divided into two categories: feature-level adaptation and instance-level adaptation. Feature-level adaptation focuses on aligning feature distributions across domains by creating a domain-agnostic latent space, while instance-level adaptation adjusts the weights of source instances to prioritize those more similar to the target domain. These approaches collectively enhance the model's ability to perform well across varied contexts and datasets \citep{deng2022domain,chunling2023adversarial}.

\textbf{Data Augmentation:} Data Augmentation involves modifying and enhancing original text data through various techniques to address data scarcity and enrich the dataset. Common methods include Easy Data Augmentation (EDA), which involves random insertion, deletion, swapping, and synonym replacement, and back translation, where text is translated to another language and then translated back to create variations. With advancements in generative adversarial networks and pretrained models, new generative-based augmentation methods have also emerged, further expanding the techniques available for enriching text data \citep{wang2024meta}.

\textbf{Adversarial Training:} Adversarial Training involves using adversarial loss methods, inspired by generative adversarial networks (GANs), to enhance domain adaptation. This approach typically includes techniques like the Domain Adversarial Neural Network (DANN), which employs a gradient reversal layer to obscure the domain discriminator and help the feature extractor learn domain-invariant representations. Another method, Adversarial Discriminative Domain Adaptation (ADDA), combines a discriminative approach with GAN loss and separate weights to reduce domain discrepancies. These methods are particularly effective for tasks like zero-shot stance detection, where they help align features across different domains \citep{chunling2023adversarial}.

\textbf{Keyphrase Generation/Extraction:} Keyphrase Generation/Extraction is the process of identifying keyphrases that accurately capture the essence or main topics of a given document, such as a research paper or news article. In the context of stance detection, this technique can be applied to generate keyphrases that are specifically related to the target of interest within the text. A widely used approach is the One2Seq model, an encoder-decoder framework that generates keyphrases sequentially in an auto-regressive fashion. This framework often utilizes pre-trained models like BART, which can be fine-tuned on datasets such as OpenKP, KP Times, and FullTextKP for generating relevant keyphrases. Despite its potential, the application of keyphrase generation specifically for target-related tasks in stance detection remains underexplored \citep{li2023new}.

\textbf{Meta Learning:} Meta Learning aims to improve a model’s ability to adapt by learning from multiple tasks, rather than focusing solely on individual tasks. This approach, inspired by transfer learning, teaches models to ``learn how to learn," which is especially beneficial for scenarios with limited data or new tasks. Model-Agnostic Meta-Learning (MAML) is a popular technique that achieves this by optimizing model parameters in two stages: first, updating the model using a specific task's data, and second, refining the model’s initial parameters based on overall performance across tasks. MAML has shown promise in various natural language processing tasks and is used here to train a model for stance detection, enhancing its capacity to quickly adjust to new and different targets \citep{wang2024meta}.

\textbf{Attention Mechanism:} The attention mechanism in stance detection is a powerful tool that enhances the model’s ability to focus on relevant parts of the input text, thereby improving the accuracy of stance classification. This mechanism allows the model to weigh different parts of the input differently, giving more importance to the words or phrases that are most indicative of the stance \citep{alturayeif2023systematic,xu2018cross}.

\section{Approaches to cross-target stance detection}

Cross-target stance detection has witnessed significant evolution and innovation over the last decade and particularly from 2016 to 2024, with researchers exploring various methods to improve model performance across diverse datasets and scenarios. In what follows we discuss existing methods to cross-target stance detection, which we group into five major types of methods: (1) Statistics-based methods, (2) Fine-tuning-based methods, (3) Prompt-tuning-based methods, (4) Knowledge-enhanced methods, and (5) Knowledge-enhanced Prompt-tuning methods. 

\begin{sidewaystable}[htbp]
    \centering
    \resizebox{\textwidth}{!}{ 
    \begin{tabular}{|p{4cm}|l|c|c|c|c|c|c|c|c|}
        \hline
        \textbf{Method} & \textbf{Model} & \textbf{Contrastive Learning} & \textbf{Domain Adaptation} & \textbf{Data Augmentation} & \textbf{Adversarial Training} & \textbf{Keyphrase Generation/Extraction} & \textbf{Graph-based} & \textbf{Meta Learning} & \textbf{Attention Mechanism} \\ \hline
        
        \multirow{6}{*}{\textbf{Statistics-based}} 
        & BiCond & & & & & & & & \\ \cline{2-10}
        & CrossNet & & & & & & & & \checkmark \\ \cline{2-10}
        & VTN & & & \checkmark & & & & & \checkmark \\ \cline{2-10}
        & TGA-Net & & & & & & & & \checkmark \\ \cline{2-10}
        & TPDG & & & & & & \checkmark & & \checkmark \\ \cline{2-10}
        & TOAD & & \checkmark & & \checkmark & & & & \checkmark \\ \hline

         \multirow{14}{*}{\textbf{Fine-tuning}} 
        & DTCL & \checkmark & & & & & & & \\ \cline{2-10}
        & GDA-CL & \checkmark & & \checkmark & \checkmark & & & & \\ \cline{2-10}
        & UTDA & & \checkmark & & & & & & \\ \cline{2-10}
        & JointCL & \checkmark & & & & & & & \\ \cline{2-10}
        & PT-HCL & \checkmark & & & & & & & \\ \cline{2-10}
        & SSCL & \checkmark & & & & & & & \\ \cline{2-10}
        & CT.AAD & & \checkmark & & \checkmark & & & & \\ \cline{2-10}
        & TSE & & & & & \checkmark & & & \\ \cline{2-10}
        & STANCE-C3 & \checkmark & \checkmark & & & & & & \\ \cline{2-10}
        & FEGCL & \checkmark & & & & \checkmark & & & \\ \cline{2-10}
        & MPCL & \checkmark & & & & & & & \\ \cline{2-10}
        & MCLDA & \checkmark & & & \checkmark & & \checkmark & \checkmark & \\ \cline{2-10}
        & MSFR & \checkmark & & & & & & & \\ \cline{2-10}
        & GLAN & & & & & & \checkmark & & \checkmark \\ \hline
        
        \multirow{11}{*}{\textbf{Prompt-based}} 
        & PET & & & & & & & & \\ \cline{2-10}
        & TAPD & & & & & & & & \\ \cline{2-10}
        & FECL & \checkmark & & \checkmark & & & & & \checkmark \\ \cline{2-10}
        & CCSD & \checkmark & & & & \checkmark & & & \\ \cline{2-10}
        & TTS & & & \checkmark & & \checkmark & & & \\ \cline{2-10}
        & Stance Reasoner & & & \checkmark & & & & & \\ \cline{2-10}
        & MTFF & \checkmark & & \checkmark & & & & \checkmark & \checkmark \\ \cline{2-10}
        & MPTT & & \checkmark & & & & & & \checkmark \\ \cline{2-10}
        & EZSD-CP & \checkmark & & & & & & & \\ \cline{2-10}
        & DS-ESD & & & & & & & & \checkmark \\ \cline{2-10}
        & EDDA & & & \checkmark & & & & & \\ \hline
        
        \multirow{6}{*}{\textbf{Knowledge-enhanced}} 
        & SEKT & & \checkmark & & & \checkmark & & \checkmark & \\ \cline{2-10}
        & CKE-Net & & & & & & \checkmark & & \checkmark \\ \cline{2-10}
        & BS-RGCN & & & & & & \checkmark & & \checkmark \\ \cline{2-10}
        & WS-BERT & & \checkmark & & & \checkmark & & & \\ \cline{2-10}
        & TarBK-BERT & & & & & \checkmark & & & \\ \cline{2-10}
        & NPS4SD & & & & & & & & \\ \cline{2-10}
        & ANEK & \checkmark & \checkmark & & \checkmark & &\checkmark & & \checkmark \\ \cline{2-10}
        & CNet-Ad & &\checkmark & &\checkmark &  &\checkmark & & \\ \cline{2-10} \hline
        
       \multirow{6}{*}{\parbox{4cm}{\centering \textbf{Knowledge-enhanced} \\ \textbf{prompt-based}}}
        & INJECT & & & & & & & & \checkmark \\ \cline{2-10}
        & \textit{COLA} & &\checkmark & & & & & & \\ \cline{2-10}
        & \textit{LKI-BART} &\checkmark & & & &\checkmark & & & \\ \cline{2-10}
        & \textit{KAI} & &\checkmark & & & & & &\checkmark \\ \cline{2-10}
        & \textit{PSDCOT} & & & & & & & &\checkmark \\ \hline
        
    \end{tabular}
    }
    \caption{Comparison of models across different techniques grouped by method}
    \label{tab:model_comparison}
\end{sidewaystable}

\subsection{Statistics-based methods:}
Statistics-based methods in cross-target stance detection primarily utilize statistical and machine learning techniques to understand and classify stances in textual data. These approaches often involve models like LSTMs, variational networks, or attention mechanisms that extract features from text and encode dependencies between the content and the target stance. They rely on traditional and statistical machine learning methods to analyze relationships between targets and the stances expressed in text.

In 2016, Augenstein et al. introduced \textbf{BiCond}, a method based on bidirectional conditional LSTM encoding, to address stance detection—a task of classifying the attitude expressed in a text towards a target. They explore a challenging scenario where the target is not always mentioned in the text and no training data is available for the test targets. The proposed BiCond method builds a representation of the tweet dependent on the target, outperforming independent tweet and target encoding. The method is evaluated on the SemEval 2016 Task 6 Twitter Stance Detection corpus, achieving second-best performance without additional training data for the test target and state-of-the-art results when augmented with weak supervision. The study utilizes several datasets: TaskA training and development data for targets like Hillary Clinton and Atheism, an unlabelled corpus of Donald Trump tweets, and automatically labeled data for weak supervision. The bidirectional conditional LSTM encoding method significantly improves F1 scores in stance detection for unseen targets, achieving 0.4901 F1 in the unseen target setting and 0.5803 F1 with weak supervision, thereby demonstrating the effectiveness of their approach in a realistic scenario where labeled data for each target is not available \citep{augenstein2016stance}.

In 2018, Xu et al. proposed \textbf{CrossNet}, a model designed for cross-target stance classification, aimed at identifying user stances on various targets by leveraging shared knowledge across related targets. CrossNet employs a self-attention mechanism to detect and utilize domain-specific aspects from a source target, enhancing generalization to a destination target. This model is structured into four layers: Embedding, Context Encoding, Aspect Attention, and Prediction. Compared to the BiCond model, CrossNet incorporates a self-attention mechanism that better captures and generalizes domain-specific information, thus improving performance in cross-target stance classification tasks. The study evaluates CrossNet using the SemEval-2016 Task 6 Twitter stance detection dataset and a collection of tweets about an Australian mining project. The SemEval dataset includes five targets: Climate Change, Feminist Movement, Hillary Clinton, Legalization of Abortion, and Donald Trump, organized into three domains: Women’s Rights, American Politics, and Environment. In the cross-target experiments, CrossNet leverages shared information across these topics to improve stance detection accuracy. For instance, it can use knowledge from stances on Climate Change to better predict stances on Hillary Clinton or the Legalization of Abortion. CrossNet demonstrates superior performance over the BiCond model, particularly in these cross-target scenarios, with an average F1-score improvement of 6.6\%. This significant performance boost highlights CrossNet's enhanced ability to generalize knowledge from one target to another, supported by both quantitative metrics and qualitative visualizations of the model's attention mechanisms \citep{xu2018cross}.

In 2019, Wei et al. introduced the \textbf{VTN} (Variational Topic Network) model for cross-target stance detection. The VTN model addresses the challenge of limited labeled data for new targets by leveraging transferable topics between a source and a destination target. The core idea is to use shared latent topics as transferable knowledge, enabling the model to generalize across different targets. This is achieved through neural variational inference, which extracts topic knowledge from unlabeled data, and adversarial training, which encourages the model to learn target-invariant representations. Compared to previous models like CrossNet, which primarily use self-attention for domain-specific feature extraction, VTN explicitly models and transfers shared topics between targets, leading to improved cross-target stance detection. The VTN model was evaluated using datasets from the SemEval-2016 Task 6 Twitter stance detection challenge, which includes targets such as Climate Change, Feminist Movement, Hillary Clinton, Legalization of Abortion, and Donald Trump. In the cross-target experiments, the model demonstrated its ability to utilize shared latent topics, such as ``equality" in the case of Feminist Movement and Legalization of Abortion, to improve stance classification on the destination target. This approach effectively bridges the gap between targets by focusing on common topics discussed within different contexts. Experimental results showed that VTN outperformed state-of-the-art methods, achieving superior cross-target stance detection accuracy by leveraging both labeled and unlabeled data to capture and transfer relevant topic knowledge \citep{wei2019modeling}.

In 2020, the \textbf{TGA-Net} (Topic-Grouped Attention Network) was introduced for zero-shot stance detection, addressing the challenge of stance classification across a wide range of topics without any training examples. The paper first presents the VAST dataset, specifically designed to encompass diverse topics with varying lexical expressions, crucial for evaluating models' ability to generalize. Unlike existing datasets with limited topics and expressions, VAST includes a broad spectrum of themes such as politics, education, and public health, facilitating a more comprehensive assessment of zero-shot and few-shot stance detection models. TGA-Net itself leverages generalized topic representations derived through unsupervised contextualized clustering, allowing the model to implicitly capture relationships between topics without human-defined rules. This approach enhances performance on challenging linguistic phenomena like sarcasm and reduces reliance on sentiment cues, which often lead to classification errors in stance detection tasks. TGA-Net's architecture comprises a contextual conditional encoding layer followed by topic-grouped attention using learned generalized topic representations. By embedding documents and topics jointly using BERT and computing generalized topic centroids through clustering, the model effectively encodes topic relationships without explicit supervision. This enables TGA-Net to outperform existing methods like BiCond and CrossNet, particularly in scenarios where there are few or no labeled examples for new topics. Experimental results on the VAST dataset demonstrate TGA-Net's superiority, achieving statistically significant improvements in macro-average F1 scores across both zero-shot and few-shot stance detection tasks, thereby validating its efficacy in handling a wide array of topics with varied lexical expressions \citep{allaway2020zero}.

In another line of work, researchers introduced the \textbf{TPDG} framework, designed to tackle the challenge of stance detection across multiple targets, including those not seen during training. The key innovation of TPDG lies in constructing heterogeneous target-adaptive pragmatics dependency graphs for each sentence relative to a given target. These graphs integrate syntactic dependency and pragmatics information derived from annotated training data and word-level analyses across various targets. Specifically, TPDG employs two types of graphs: in-target graphs, which capture pragmatics dependencies specific to each target, and cross-target graphs, which enhance the adaptability of words across all targets \citep{liang2021target}. The integration of these heterogeneous graphs is achieved through a graph-aware model utilizing interactive Graph Convolutional Network (GCN) blocks. These blocks allow the model to dynamically adjust and learn from both target-specific and target-independent contextual graph representations. This approach improves the model’s ability to understand and adapt stance expressions for unseen targets by leveraging knowledge from the constructed graphs. Experimental results on benchmark datasets, SemEval-2016 Task 6 and Wt-wt, show that the TPDG model outperforms existing state-of-the-art methods in cross-target stance detection. For SemEval-2016, evaluation was performed using the mean value of the Macro F1-score for the 'favor' and 'against' classes, as well as the average of both micro-averaged and macro-averaged F1 scores to address target imbalance. For the Wt-wt dataset, the Macro F1-score for all labels was used to assess performance across all targets. These results highlight the effectiveness of incorporating external knowledge, including both target-specific and target-independent pragmatics dependencies, into the model. The TPDG framework not only enhances stance detection accuracy but also improves interpretability, aligning with knowledge-enhanced methods in natural language processing research.

Allaway et al. introduced the \textbf{TOAD} (TOpic ADversarial network) model for zero-shot stance detection, which employs adversarial learning to generate topic-invariant representations, enabling it to generalize across various topics on Twitter. The model uses a bidirectional conditional encoding (BiCond) architecture to create topic-specific document representations, which are then transformed through a linear layer to become topic-invariant. An adversary, trained simultaneously with the stance classifier, helps ensure that these representations do not encode topic-specific information, thereby improving the model's ability to handle unseen topics. The TOAD model incorporates a stance classifier and a topic discriminator, where the former predicts stance labels and the latter attempts to identify the topic from the transformed representations. The adversarial training process minimizes the classifier’s error while maximizing the adversary’s error, effectively promoting the creation of topic-invariant features \citep{allaway2021adversarial}. The dataset used in this study is the SemEval2016 Task 6 (SemT6), which includes six topics: Donald Trump (DT), Hillary Clinton (HC), Feminist Movement (FM), Legalization of Abortion (LA), Climate Change (CC), and Atheism (A). The TOAD model is trained on five of these topics and tested on the remaining one in a cross-target setting, where each topic in turn is used as the zero-shot test topic. The model was also benchmarked on two additional topics not previously used in zero-shot settings, namely Atheism and Climate Change. The results show that TOAD achieves state-of-the-art performance on the majority of topics, including DT and FM, and performs comparably to BERT on others, while being significantly more computationally efficient. The analysis indicates that TOAD's adversarial learning approach is effective in generating robust, topic-invariant representations, leading to superior performance in zero-shot stance detection scenarios. They reported \( F_{\text{avg}} \) as the evaluation metric, the average of the F1 scores on the 'pro' and 'con' stances.

\subsection{Fine-tuning based methods:}

Fine-tuning-based methods involve adapting pre-trained language models (like BERT or GPT) to specific stance detection tasks by further training them on domain-specific datasets. This approach leverages the extensive pre-existing knowledge in these models, refining them to recognize and classify stances by using labeled data related to the target topics. It is particularly effective in scenarios where the model must generalize across different but related targets by fine-tuning with minimal data.

\textbf{BertEmb} uses BERT embeddings with a multi-layer perceptron (MLP) for stance detection (SD) and evidence retrieval (ER), effectively capturing contextual information from complex, unstructured texts like news articles. It highlights the power of transfer learning and pre-trained models for nuanced stance detection, especially in news reports, addressing gaps left by previous research focused mainly on user-generated content. The model is tested on the STANDER dataset, containing 3,291 expert-annotated news articles about US healthcare mergers. It is evaluated with macro-averaged precision, recall, and F1 for SD, and precision and recall for ER. BertEmb achieves the highest scores for SD in cross-target and zero-shot settings, though it still falls short of human performance, suggesting a need for further improvements in multi-task learning and domain adaptation \citep{reimers2019sentence,conforti2020stander}.

In 2022, the \textbf{DTCL} model was proposed for zero-shot and few-shot stance detection (ZFSD) using BERT to encode document-target pairs. It addresses the challenge of stance detection with limited or no training data by introducing a discrete latent topic variable that models relationships between seen and unseen targets in a latent space, enabling automatic topic clustering. To improve generalization, DTCL uses supervised contrastive learning to learn target-invariant features, overcoming limitations of previous models that relied on large unlabeled datasets or poorly modeled target relationships \citep{liu2022connecting}. DTCL is evaluated on the VAST dataset with the macro-averaged F1 score as the metric. It outperforms baselines like TGA-Net, which suffers from training-representation mismatches, and CKE-Net, which relies on external knowledge. Without such dependencies, DTCL, powered by BERT, demonstrates superior performance in challenging scenarios like handling neutral labels, sarcasm, and multiple targets, showing its robustness in modeling document-target relationships.

The \textbf{GDA-CL} model tackles zero-shot stance detection (ZSSD) by generating high-quality synthetic training data for unseen targets using a combination of generative adversarial networks (GANs) and hybrid contrastive learning. Unlike methods relying on external knowledge, GDA-CL creates training samples in the same embedding space as real ones, improving knowledge transfer and stance detection performance. It leverages GPT-2 as the generator, RoBERTa as the discriminator, and BERT as the classifier within a GAN framework, with an MLP for contrastive learning. Evaluated on VAST and SemT6 datasets, GDA-CL achieves state-of-the-art results across most topics by using data augmentation and robust contrastive learning \citep{li2022generative}. In the cross-target setting, GDA-CL consistently outperforms baselines like TOAD, BERT-GCN, and CrossNet by generating high-quality data for unseen targets. Although it excels with shorter, coherent texts, the model struggles with longer texts, showing less improvement on the VAST dataset. This highlights its limitation in capturing complex semantics in longer texts, suggesting an area for future refinement.

Deng et al. introduced \textbf{UTDA} (Unified Target-aware Domain Adaptation) for cross-target stance detection using transformer-based language models (TLMs) to adapt to new targets on social media. Unlike traditional methods relying on manually selected target pairs, UTDA uses unsupervised feature disentanglement and instance weighting to automatically identify and adapt to target relations. Evaluated on SemEval2016-T6 and COVID-19-STANCE datasets under a leave-one-out setting, UTDA outperforms baselines like TOAD, DANN, and MOE, improving over vanilla BERT by 7.2\% and surpassing MOE by 8.9\% in average F1-score, particularly excelling with targets having no predefined relations \citep{deng2022domain}. The study validates UTDA's effectiveness by leveraging TLMs and enhancing model adaptability without predefined target pairs. Its strong performance on both datasets shows its ability to handle dynamic and diverse targets, offering a scalable solution for adapting to emerging topics and sentiments on social media.

The \textbf{JointCL} model introduces Joint Contrastive Learning for zero-shot stance detection (ZSSD), focusing on detecting stances for unseen targets \citep{liang2022jointcl}. It integrates two strategies: stance contrastive learning, which enhances stance feature generalization by clustering similar instances within stance classes, and target-aware prototypical graph contrastive learning, which uses prototypical graphs to model relationships between known and unseen targets, transferring stance information across related targets. Evaluated on the VAST, SEM16, and WT-WT datasets, JointCL outperforms baselines like BERT and SEKT in cross-target scenarios, such as training on one target and testing on a related unseen target (e.g., HC to DT, FM to LA) on the SEM16 dataset. This demonstrates JointCL's effectiveness in adapting to unseen targets by leveraging context-aware and target-aware perspectives, achieving state-of-the-art performance in ZSSD where traditional supervised models falter due to a lack of labeled data.

The \textbf{PT-HCL} framework enhances zero-shot stance detection (ZSSD) by distinguishing between target-invariant and target-specific stance features \citep{liang2022zero}. It uses a two-fold approach: first, self-supervised learning with a pretext task to augment training data with masked instances based on target-related words, and second, a hierarchical contrastive learning framework that integrates stance labels and feature types. In cross-target stance detection, PT-HCL shows significant improvements over baselines like BiCond and CrossNet, particularly in scenarios such as ``HC→DT" and ``DT→HC." The model achieves higher accuracy and F1 scores, demonstrating its robust ability to generalize stance detection across unseen targets and advance the state-of-the-art in ZSSD. This makes PT-HCL a promising approach for detecting stances towards previously unseen targets in real-world applications.

The \textbf{SSCL} (Sentiment-Stance Contrastive Learning) model addresses zero-shot stance detection (ZSSD) by focusing on cross-target stance detection. It utilizes contrastive learning to extract target-invariant features from texts, analyzing both sentiment and stance dimensions. The model groups texts accordingly and employs a supervised contrastive learning approach to capture transferable features, which are then combined with target-specific semantic information to improve stance detection for unseen targets. Evaluated on three benchmark datasets, including SEM16, SSCL achieved state-of-the-art performance in cross-target stance detection. The evaluation used the mean of micro-averaged and macro-averaged F1 scores for Favor and Against, addressing data imbalance issues \citep{zou2022zero}. However, SSCL has some limitations. It focuses on extracting features from sentiment and stance dimensions, potentially neglecting finer details like syntactic structure and perspective that could enhance stance detection. Additionally, its reliance on labeled sentiment and stance data limits the use of large unlabeled datasets during testing, which may impact its generalization capabilities.

Pavan et al. introduced the \textbf{CT.AAD} model for cross-target stance classification, adapting BERT with adversarial learning and knowledge distillation to enhance stance detection on unseen targets. Using the UstanceBR r1 dataset, which includes Portuguese tweets on six topics like political figures and COVID-19, CT.AAD was benchmarked against single-target models and general LSTM-BERT approaches. Results show CT.AAD performs better than other cross-target methods but remains behind single-target models, especially in polarized contexts. Evaluation metrics focus on F1 scores, accuracy, and loss \citep{pavan2022cross}. In 2023, the Multi-Perspective Contrastive Learning Framework emerged for cross-target stance detection (CTSD) and zero-shot stance detection (ZSSD). Leveraging both labeled and unlabeled data with BERT base, the model was evaluated on the SemEval-2016 dataset. It achieved comparable results to strong baselines like SKET and TAPD in tasks such as FM→LA and HC→DT, though its performance may be limited by the number of targets in CTSD. This framework improves stance detection by integrating unlabeled texts, addressing the issues of limited labeled data and inaccuracies in external knowledge \citep{jiang2023zero}.

Li et al. introduced the \textbf{TSE} (Target-Stance Extraction) task for in-target and cross-target stance detection, focusing on automatically extracting target and stance pairs from text, especially when targets are implicit. The approach uses a two-stage framework: target identification and stance detection. Target identification includes Target Classification with classifiers like BiLSTM and BERT, and Target Generation with a fine-tuned BART model to generate keyphrases mapped to predefined targets. For stance detection, a multi-task learning approach is employed, with target prediction as an auxiliary task to enhance focus on target-related features. The model is evaluated on a combined dataset from four stance datasets and a new zero-shot dataset with unseen targets such as ``Creationism" and ``MeToo Movement." Metrics include F1 score and accuracy for target-stance extraction, and micro- and macro-F1 for target identification and stance detection. TGA-Net, utilizing topic-grouped attention, outperforms other models in zero-shot settings, demonstrating the TSE framework’s effectiveness for applications with undefined target sets \citep{li2023new}.

Kim et al. introduced \textbf{STANCE-C3}, a method for stance detection that combines domain counterfactual generation with contrastive learning. The model features a T5-based domain counterfactual generator, which adapts source domain samples to target domains, and a BERT-based stance classifier that incorporates these counterfactuals. The classifier uses contrastive learning to distinguish stance types by minimizing distances between positive pairs and maximizing those between negative pairs, enhancing generalization across targets \citep{kim2023stance}. Evaluated on the CoVaxNet dataset (COVID-19 vaccination) and the COVID-19-Stance dataset (varied stance targets), STANCE-C3 showed strong performance in few-shot and zero-shot scenarios. The model's effectiveness was measured using classification accuracy, AUC, and contrastive loss, demonstrating improvements in stance detection across single-domain, cross-domain, and cross-target settings.

Zhao et al. introduced \textbf{FEGCL} (Feature Enhancement via Graph Contrastive Learning) for stance detection on unseen targets, addressing both zero-shot and cross-target scenarios. FEGCL uses a dual-view graph construction to capture target-invariant features from syntactic and semantic perspectives. It combines unsupervised graph contrastive learning with an interactive GCN to enhance stance detection capabilities for unseen targets while maintaining global semantic consistency \citep{zhao2023unified}. Evaluated on the VAST dataset (zero-shot) and the WT-WT dataset (cross-target), FEGCL shows superior performance, with a higher F1-Score than baseline methods. It demonstrates effective feature extraction and stance detection, achieving robust results in both zero-shot and cross-target scenarios, and excels in generalizing to new, unseen targets through graph-based contrastive learning.

Zhao et al. focused on zero-shot stance detection (ZSSD) with a specific emphasis on cross-target stance detection using the Chinese C-STANCE dataset. The paper tackles two subtasks: target-based ZSSD, testing models on unseen targets, and domain-based ZSSD, evaluating models on targets from new domains. The dataset includes 48,126 annotated text-target pairs from Sina Weibo, covering diverse controversial topics \citep{zhao2023c}. The evaluation metric used is the macro-averaged F1 score. The best-performing model, RoBERTa, achieved an F1 score of 78.5\% on target-based ZSSD, highlighting room for improvement, particularly in handling varied target types. The study establishes baseline results with several deep learning models, including BERT, RoBERTa, and XLNet, within a zero-shot learning framework where models encounter test set targets only during evaluation.

\textbf{MPCL} or \textbf{Mult-CL} (Multi-Perspective Contrastive Learning) addresses zero-shot stance detection (ZSSD) by leveraging both labeled and unlabeled data to enhance target representations. The framework includes BERT for text encoding, target-oriented contrastive learning (Target-CL) for refining target representations, and label-oriented contrastive learning (Label-CL) for stance features across different labels. This approach improves upon previous methods by handling noisy schema-linking and lacking target-specific information. MPCL achieves state-of-the-art results on datasets such as SemEval-2016, WT-WT, and VAST, showing notable improvements over baselines in SemEval-2016 and WT-WT, and comparable performance on VAST \citep{jiang2023zero}. In cross-target stance detection (CTSD), MPCL competes well against models like SKET and TPDG and performs better than PT-HCL on most target pairs, except FM→LA and DT→HC. Evaluation on SemEval-2016 shows MPCL’s stability across cross-target tasks, though it struggles with fewer training targets, affecting its overall effectiveness. The model’s robust performance is attributed to its ability to balance target-specific and independent features through contrastive learning.

\textbf{CT-TN} (Cross-Target Text-Net), proposed by Jamadi Khiabani et al., combines multimodal embeddings from RoBERTa for text and network-based embeddings from social interactions using PecanPy, a variant of Node2Vec. The model processes text and network features in parallel and uses a majority voting mechanism for stance prediction. CT-TN outperforms models like TGA-Net, CrossNet, and RoBERTa by incorporating social network interactions, achieving 11\% to 21\% improvement in macro-averaged F1 scores over baseline models \citep{khiabani2023few}. In cross-target stance detection with the P-Stance dataset, CT-TN excels, particularly with 300+ examples per target, showing up to 21\% higher F1 scores compared to other models. It performs well even with contrasting ideologies, such as Trump and Sanders, and benefits from network features, enhancing accuracy across various source-destination pairs. However, performance diminishes with fewer examples, but the inclusion of network features still makes CT-TN a robust choice for diverse and cross-target stance detection tasks.

Tacchi et al. present the Ego Network Model (ENM) and the Signed Ego Network Model (SENM) for cross-target stance detection (called \textbf{Stance+Ego }model), leveraging social network data to enhance stance prediction. The methodology involves transforming both textual and social network data into vector embeddings using node2vec for graph-based features and RoBERTa for text-based predictions. The ENM and SENM are evaluated against the CT-TN baseline and RoBERTa on the P-Stance dataset, which includes 21,574 tweets about political figures, using a few-shot learning approach. Specifically, the experiments use 1,000 source-target data points with varying numbers of destination-target texts, ranging from 100 to 400 shots. The results reveal that while the ENM and SENM slightly lag behind the CT-TN model, they perform comparably with less data, demonstrating their efficacy in stance detection. Notably, the outer circles of the ENM, which capture weaker but more numerous social ties, prove to be more informative for stance prediction than the inner circles. This suggests that less intimate connections have a significant impact on stance detection, and the ENM offers a viable alternative to more data-intensive methods. It is important to note that they used Macro F1 score as the evaluation metric \citep{tacchi2024applying}.

\textbf{MCLDA} (Meta-Contrastive Learning for Data Augmentation), proposed by Wang et al., addresses zero-shot stance detection (ZSSD) by mitigating data scarcity and improving model generalization. It employs two main strategies: generating target keyphrases using BART fine-tuned on KPTimes, and applying meta-contrastive learning by framing stance detection as a text entailment task, trained with MAML and contrastive learning \citep{wang2024meta}. MCLDA improves over previous methods by generating keyphrases for texts lacking explicit targets, common in short social media posts, and stabilizes performance across targets through meta-learning. It was evaluated on VAST, SemEval-2016 (SEM16), and WT-WT datasets, using metrics like Macro F1 and average F1 scores for stance detection. The model shows superior performance in cross-target scenarios, particularly with target pairs like FM-LA and DT-HC, due to its ability to leverage target correlations. In zero-shot stance detection, MCLDA competes well against models such as BiCond and CrossNet, and surpasses BERT-GCN, especially in scenarios without external knowledge. Its cross-target performance is notably strong, benefiting from learned correlations between known and unknown targets.

The \textbf{MSFR} (Multi-aspect Semantic Feature Representation) model enhances zero-shot stance detection by using a hierarchical contrastive learning framework with two components: inter-aspect contrastive learning and intra-aspect contrastive learning. Inter-aspect learning aligns semantic features across target domains, crucial for knowledge transfer, while intra-aspect learning captures fine-grained attribute-level features within aspects, improving generalization. This approach addresses the limitations of traditional methods that rely on coarse global features \citep{zhao2024msfr}. Evaluated on the VAST and WT-WT datasets, MSFR consistently outperforms baseline models, including BiLSTM, adversarial learning, and graph neural networks. It shows superior performance compared to the TOAD model, particularly in unbalanced data scenarios, and significantly surpasses PT-HCL in the WT-WT dataset. The model’s fine-grained feature extraction and BERT-based representation contribute to its effectiveness. An ablation study confirms the importance of both contrastive learning components, highlighting MSFR’s robust performance and detailed approach to feature representation and knowledge transfer in zero-shot stance detection.

The \textbf{GLAN} model incorporates three key components to enhance stance detection: a global attention layer, a local convolutional layer, and a structural layer. The global attention layer uses BERT’s pre-trained embeddings to capture long-range dependencies and integrate contextually relevant information from distant parts of a conversation. The local convolutional layer applies CNNs to detect short-range nuances and subtleties within smaller dialogue segments. The structural layer employs Graph Convolutional Networks (GCNs) to analyze the relationships between comments, incorporating structural information from the conversation history. Evaluated using the Favg metric on the MT-CSD dataset, which focuses on conversational stance detection across various topics and depths, GLAN demonstrates superior performance compared to baseline models like BERT, GCN, and CNN-based approaches. It excels in domain-specific scenarios, outperforming models such as CrossNet, KEPrompt, BERT, and TTS when the training and testing targets are within the same domain. However, in cross-domain experiments, the TTS model shows better performance, indicating its effectiveness in handling topic similarities across different domains, while GLAN proves more robust in domain-specific tasks.

\subsection{Prompt-tuning based methods:}

Prompt-tuning has been applied to various natural language processing tasks such as text classification, natural language understanding, and sentiment analysis. This approach uses task-specific prompts or templates to guide language models in understanding and classifying text, eliciting relevant information from pre-trained models with minimal additional training data. By framing tasks as fill-in-the-blank or similar cloze tasks, prompt-tuning allows models to generalize stances across various targets, leveraging the inherent linguistic patterns recognized by the language models. A key component of prompt-tuning is the verbalizer, which significantly influences its effectiveness. Verbalizers can be classified into two types: (1) human-designed verbalizers, which rely on personal expertise and often lack comprehensive coverage, as seen in Schick et al.'s manual definition of label words for text classification; and (2) automatic verbalizers, which use automatic searching methods to identify more effective verbalizers \citep{huang2023knowledge}.

In 2020, Schick et al. introduced \textbf{PET}, a pattern-exploiting training method that revolutionized language model training by utilizing cloze-style phrases to assign soft labels. This semi-supervised approach significantly enhances the understanding and performance of language models in tasks such as stance detection. PET has been successfully applied across various datasets, including Yelp Reviews and AG’s News, demonstrating its effectiveness in improving model robustness and accuracy even with limited labeled data \citep{schick2020exploiting}. Specifically, PET excels in cross-target stance detection by leveraging task-specific prompts that encapsulate nuanced linguistic cues. By integrating natural language patterns into training, PET enables pretrained language models to generalize more effectively across different targets within and across domains. Experimental results consistently show PET outperforming traditional fine-tuning methods, highlighting its capability to enhance adaptability and performance in challenging low-resource settings. This approach not only advances the field of natural language processing but also paves the way for more sophisticated AI applications that require nuanced understanding of textual data.

The \textbf{TAPD} (Target-Aware Prompt Distillation) model addresses few-shot learning challenges in stance detection by leveraging pre-trained language models (PLMs) and prompt-based fine-tuning. Unlike traditional methods with fixed prompts, TAPD adapts prompts to be target-aware, using a novel verbalizer that maps stance labels to continuous vectors for improved contextual understanding. This approach distills information from multiple prompts to handle diverse expressions across targets. Evaluated on SemEval-2016 Task 6 Sub-task A and the UKP dataset, TAPD shows superior performance in both full-data and few-shot scenarios. Specifically, in cross-target stance detection on the SemEval-2016 dataset, TAPD outperforms methods like SKET and TPDG across various target pairs, such as Hillary Clinton vs. Donald Trump and Legalization of Abortion vs. Feminist Movement. TAPD's effective use of prompt-based fine-tuning and multi-prompt distillation demonstrates its potential for adaptive stance detection in diverse topics and contexts \citep{jiang2022few}.

Zhao et al. introduced \textbf{FECL} (Feature Enhanced Zero-shot Stance Detection Model via Contrastive Learning) to address zero-shot stance detection challenges. FECL captures target-invariant features through syntactic patterns and combines them with target-specific semantic features. It uses data augmentation by masking topic words and employs contrastive learning to extract transferable syntactic features. Evaluations on datasets such as SEM16 and COVID-19 demonstrate that FECL outperforms baseline models in zero-shot, few-shot, and cross-target scenarios, with performance measured by the Macro-averaged F1 score. An ablation study on the COVID-19 dataset shows that random masking and removing the contrastive learning loss significantly reduce performance, highlighting their importance. Additional analysis with alignment and uniformity metrics, and T-SNE visualizations, confirms that contrastive learning enhances the model’s generalization capabilities. FECL achieves improved alignment and uniformity, facilitating better performance in zero-shot and cross-target tasks, and shows potential for broader application in stance detection across different domains \citep{zhao2022zero}.

Zhang et al. proposed a novel approach named \textbf{CCSD} (Cross-lingual Cross-target Stance Detection) using a dual knowledge distillation framework. This methodology includes a cross-target teacher model that learns and generalizes target category representations. The proposed Cross-lingual Cross-target Stance Detection (CCSD) method utilizes a dual knowledge distillation framework. It includes a cross-lingual teacher model, prompt-tuned with cross-lingual templates, and a cross-target teacher model that focuses on learning fine-grained target knowledge. This dual approach facilitates the transfer of language-related and target-oriented knowledge from a high-resource source language to a low-resource target language \citep{zhang2023cross}. To tackle the issues of target inconsistency, the cross-target teacher model aggregates semantically correlated target representations to extract category information. This is refined using category-oriented contrastive learning, enhancing the model's ability to generalize to unseen targets. The cross-lingual teacher serves as the initial encoder, helping to mitigate language differences during the distillation process. The combined output from both teachers provides pseudo-labels for the student model, facilitating knowledge transfer despite varying target distributions and the absence of labeled data. The methodology is validated on multilingual datasets with diverse target settings, demonstrating superior performance compared to existing methods, particularly in handling unlabeled data and unseen targets. However, challenges remain in achieving optimal performance when the nature of target expressions varies significantly across datasets.

The \textbf{TTS} (Target-based Teacher-Student) framework addresses zero-shot stance detection by expanding the training set with targets generated from an unlabeled dataset. A keyphrase generation model is used to create additional targets, and a detailed analysis reveals that 78\% of these targets are relevant to their texts. The teacher model, which leverages large language models like BERT, effectively assigns stance labels, achieving 81\% consistency in the augmented data. This setup enhances stance detection performance without requiring extensive human-annotated datasets \citep{li2023tts}. For the open-world zero-shot stance detection (ZSSD) task, the TTS framework introduces a novel approach by formulating stance detection as a Natural Language Inference (NLI) task. A synthetic training set is generated, and the BART model, pre-trained on the MNLI dataset, predicts stance labels by applying the prompt template “I am in favor of [target] !”. This reformation allows the task to be viewed through NLI’s entailment, contradiction, or neutrality labels. The method shows significant improvements, with an 8.9\% increase in macro-averaged F1 scores over previous models on VAST dataset, demonstrating its effectiveness in scenarios lacking annotated targets and labels. This approach, which includes prompt tuning techniques, not only improves zero-shot stance detection but also sets a new benchmark for future research in target-based data augmentation for stance detection tasks. 

The \textbf{Stance Reasoner} architecture leverages large pre-trained language models (PLMs) like LLaMA and Vicuna for its zero-shot stance detection tasks. These models are selected for their strong in-context learning abilities and are used in combination with the chain-of-thought (CoT) approach to generate intermediate reasoning steps that support stance predictions. Stance Reasoner also employs self-consistency, where multiple outputs are generated and a majority vote is taken to improve prediction accuracy and robustness \citep{taranukhin2024stance}.

In the zero-shot setting, Stance Reasoner surpasses several baseline models, including fully supervised methods such as BERT-GCN, and knowledge-infused models like BERT-based models fine-tuned for stance detection. Stance Reasoner consistently outperforms these baselines on datasets like SemEval-2016 Task 6a, WT-WT, and the COVID-19 Stance dataset, achieving better generalization across unseen targets by incorporating reasoning and background knowledge rather than relying solely on training data patterns.

\textbf{MTFF} (Multi-perspective Transferable Feature Fusion) is designed for zero-shot stance detection, enhancing adaptability to new contexts. It incorporates target-keyword masking for data augmentation, instance-wise contrastive learning (instance-CL) for refined meta-feature identification, and an attention mechanism for effective feature fusion. Using large language models like BERT and RoBERTa for initial feature extraction, MTFF employs pattern-based approaches and a multi-head attention mechanism to improve stance classification accuracy. Evaluations on datasets such as VAST, ProCon, WT-WT, and P-Stance show that MTFF surpasses baseline models, including RoBERTa and BERT, demonstrating superior accuracy and robustness in zero-shot tasks \citep{zhao2024zero}. MTFF excels in zero-shot scenarios by generalizing to new domains through unsupervised clustering and instance-wise contrastive learning. Performance metrics, including accuracy and F1 scores, highlight significant improvements over traditional methods that rely on global text features. Results from VAST and ProCon confirm MTFF's effectiveness in transferring knowledge and detecting stances across diverse contexts, outperforming previous models like GPT-3 and conventional frameworks.

The \textbf{MPPT} (Multi-Perspective Prompt-Tuning) model for Cross-Target Stance Detection (CTSD) utilizes large language models (LLMs) within its components to enhance performance. Specifically, the model incorporates BERT-base uncased as the underlying language model for its prompt-tuning framework (MultiPLN). Additionally, GPT-3.5-0301 is employed in the two-stage instruct-based chain-of-thought method (TsCoT) to generate natural language explanations (NLEs) from multiple perspectives. These LLMs are integral to MPPT, enabling it to effectively handle the complexities of informal text structures and implicit expressions, thus facilitating superior knowledge transfer across domains compared to other methods \citep{ding2024cross}. The MPPT model was tested on the SEM16 and VAST datasets, focusing on cross-target and zero-shot scenarios. Evaluated using the Macro F1 score, the results showed that MPPT outperformed state-of-the-art models such as PT-HCL and KE-PROMPT, with a 13\% average improvement in CTSD tasks. Moreover, MPPT outperforms the knowledge-enhanced model TarBK, highlighting that leveraging analysis perspectives to bridge knowledge gaps is more effective than depending on structural background knowledge. In the cross-target experiments on SEM16, MPPT demonstrated significant performance gains across all tasks, effectively transferring domain-invariant knowledge. Moreover, in zero-shot stance detection (ZSSD) on the VAST dataset, MPPT showed strong generalization capabilities, surpassing other models in all zero-shot conditions.

\textbf{EZSD-CP} is designed to tackle both zero-shot and cross-target stance detection by incorporating a gated multilayer perceptron (gMLP) and integrating prompt learning with contrastive learning. The gMLP module enhances the connection between prompts and instances by dynamically adjusting the influence of prompts based on semantic context, which is crucial for detecting stance across different targets. In this method, BERT and RoBERTa are used as the core pretrained language models (PLMs) for generating word embeddings and semantic representations. Unlike traditional models that rely on external knowledge (as seen in models like CKE-Net) or struggle with intrinsic data, EZSD-CP effectively utilizes these advanced strategies to improve generalization and accuracy in cross-target scenarios, where the relationship between known and unknown targets is leveraged for better performance \citep{yao2024enhancing}. The performance of EZSD-CP was evaluated on the VAST and SemEval-2016 (SEM16) datasets. For cross-target stance detection, EZSD-CP was evaluated on the SemEval-2016 (SEM16) dataset, where the model's ability to generalize across different targets was rigorously tested. Metrics such as Macro F1 and accuracy were used to assess the model's performance. EZSD-CP outperformed baseline models, demonstrating superior cross-target generalization by effectively capturing the relationships between various stance targets. The model achieved notable results, surpassing the state-of-the-art JointCL model in cross-target settings, indicating that the integration of gMLP and contrastive learning significantly enhances its ability to adapt to new and unseen targets.

The work by Motyka and Piasecki presented an overview of some state-of-the-art methods in target-phrase zero-shot stance detection, emphasizing the challenges and advancements in this field. The model architecture features a novel modification of prompt-based approaches for training encoder transformers, achieving results comparable to large language models (LLMs) but with significantly fewer parameters. Experimental results demonstrated that their prompting-based methods outperformed traditional fine-tuning approaches, particularly in zero-shot settings. Special attention is given to evaluating the effectiveness of various prompts, revealing that more complex, non-annotation-task-based prompts tend to yield better performance, especially for controversial topics \citep{motyka2024target}. The experiments utilized the SemEval-2016 Task 6 and VAST datasets, with metrics like the average F1 score for favor and against classes, and the average F1 score for all three classes including zero-shot parts. Results showed that the prompting approach with RoBERTa-large achieved state-of-the-art performance on these datasets. While LLMs like GPT-3.5-turbo and FLAN-UL2 demonstrated strong zero-shot capabilities, especially on SemEval-2016, their performance was variable depending on the prompts used. In contrast, the proposed prompt-based methods consistently outperformed or matched LLMs in efficiency and effectiveness, highlighting the importance of high-quality, well-designed prompts. The study also underscored the need for better quality training data and the challenges posed by true neutral samples in datasets like VAST.

The promot-based \textbf{DS-ESD} framework developed by Ding et al. addresses cross-target stance detection using a novel distantly supervised method. Its architecture incorporates an instruction-based Chain-of-Thought (CoT) method with a very large language model (VLLM) such as GPT-3.5, a generative network that maps inputs to explanations, and a BERT-based stance classifier trained on these explanations. This design minimizes the need for extensive manually labeled data and enables effective stance detection through distant supervision. Additionally, DS-ESD features an adaptive training mechanism, including curriculum learning and label rectification, which enhances the model's performance in handling noisy labels. Experimental results on datasets like SemEval-2016 Task 6, COVID-19, and VAST revealed that DS-ESD significantly outperformed baseline methods, with an average improvement of 12.85\% in F1 score over top statistical methods and 12.05\% over fine-tuning methods. The model also demonstrated robustness in zero-shot scenarios, proving its effectiveness in stance detection even with limited annotated data \citep{ding2024distantly}.

The proposed \textbf{EDDA} (Encoder-Decoder Data Augmentation) framework proposed by Ding et al. addresses the limitations of existing data augmentation techniques in Zero-Shot Stance Detection (ZSSD). EDDA employs a unique methodology where an encoder uses large language models (LLMs) with chain-of-thought prompting to generate target-specific if-then rationales, thus creating logical relationships between the text and the targets. This helps to maintain semantic coherence while the decoder generates new samples through a semantic correlation word replacement strategy, increasing syntactic diversity. Additionally, EDDA incorporates a rationale-enhanced network (REN) to effectively utilize the augmented data, bolstering the model's ability to generalize to unseen targets \citep{ding2024edda}. The methodology was validated on SEM16 dataset for cross-target scenario. For cross-target evaluation, the framework was tested on the SemEval-2016 Task 6 (SEM16) dataset. The evaluation metric employed was the Macro-averaged F1 score across the “Favor” and “Against” classes. This metric was used to measure the framework's effectiveness in handling unseen targets, demonstrating that EDDA significantly outperforms traditional methods by producing more generalized and diverse samples. The results demonstrated that EDDA-generated data has greater diversity and relevance, leading to improved performance across various ZSSD models, including BiLSTM, CrossNet, TGA-Net, Bert-Joint, and JointCL. This highlights EDDA's robustness in handling new, unseen topics or entities, making it a potent tool for stance detection tasks.

The \textbf{COLA} (Collaborative rOle-infused LLM-based Agents) framework, built on the GPT-3.5 Turbo model from OpenAI, leverages its high performance and cost-effectiveness through the OpenAI API. The architecture includes a multidimensional text analysis stage with three expert agents—Linguistic Expert, Domain Specialist, and Social Media Veteran—that analyze text from various perspectives. This is followed by a reasoning-enhanced debating stage where a Judger agent synthesizes the insights from the experts to determine the stance. Prompts guide these roles, ensuring systematic analysis and reproducibility. COLA’s design improves upon existing methods by offering robust zero-shot stance detection without additional model training \citep{lan2024stance}. COLA’s effectiveness is evaluated using Favg (average F1 score for Favor and Against) for the SEM16 and P-Stance datasets, and Macro-F1 score for the VAST dataset. It outperforms both zero-shot and in-target stance detection methods, showing high accuracy and effectiveness across these datasets. The framework also matches state-of-the-art baselines in related text classification tasks, demonstrating its versatility and practical application. Additionally, COLA provides clear and rational explanations for its decisions, enhancing its usability and trustworthiness. Future work will focus on incorporating real-time knowledge updates to improve analysis of current events and extend its capabilities in web and social media text analysis.

\subsection{Knowledge-enhanced methods:}
Knowledge-enhanced methods integrate external knowledge sources, such as semantic lexicons, commonsense knowledge, or domain-specific databases, into the stance detection process. These approaches use knowledge graphs or databases to provide context and enhance the understanding of the text, enabling the model to make more informed decisions about stances. This is particularly useful in zero-shot or few-shot learning scenarios, where labeled data for specific targets may be scarce or unavailable.

The \textbf{SEKT} (Semantic-Emotion Knowledge Transferring) model proposed by Zhang et al. addresses cross-target stance detection by leveraging external semantic and emotion lexicons to enhance representation learning. SEKT constructs a semantic-emotion heterogeneous graph from these lexicons, utilizing graph convolutional networks (GCN) to capture multi-hop semantic connections between words and emotion tags. This graph-based approach enriches the model's understanding of stance by integrating domain-specific knowledge, facilitating effective knowledge transfer across different targets. Moreover, SEKT extends the bidirectional LSTM (BiLSTM) with a novel knowledge-aware memory unit (KAMU), enabling the model to integrate external knowledge seamlessly into the stance classification process \citep{zhang2020enhancing}. The researchers evaluated SEKT on a comprehensive dataset derived from SemEval-2016 Task 6, a benchmark for stance detection in social media texts. Their experiments demonstrated SEKT's superior performance over state-of-the-art methods, including traditional BiLSTM and BERT models, as well as other graph-based approaches like CrossNet. SEKT consistently outperformed these baselines across multiple cross-target stance detection tasks, showcasing its effectiveness in handling short, informal texts and implicit stance expressions. This success underscores SEKT's role in advancing stance detection research, particularly in scenarios where labeled data is scarce or when targeting diverse subjects across different domains.

In another line of work called \textbf{CKE-Net}, the authors address the challenging problem of zero-shot and few-shot stance detection, focusing on scenarios where very limited or no annotated training data is available for new topics. Traditional data-driven approaches struggle in such settings due to their dependency on large annotated datasets. To overcome these limitations, the authors propose a novel approach that integrates commonsense relational knowledge from ConceptNet into the stance detection process. They leverage a relational subgraph extracted from ConceptNet, capturing semantic relationships between concepts to facilitate reasoning about stances across diverse topics. This approach not only enhances the model's generalization capabilities but also mitigates the need for extensive annotated data \citep{liu2021enhancing}. The methodology involves encoding textual inputs using BERT and enriching them with relational knowledge via a Graph Convolution Network (GCN), specifically CompGCN, tailored for relational graphs. This integration enables the model to effectively combine structural and semantic information from the knowledge graph with contextual information from the text. Experimental results on their newly introduced dataset, VAried Stance Topics (VAST), demonstrate significant performance improvements over state-of-the-art methods. Their model achieves superior performance across various evaluation metrics, showcasing its effectiveness in handling both zero-shot and few-shot stance detection tasks. This underscores the critical role of external commonsense relational knowledge in enhancing the robustness and applicability of stance detection models in challenging, data-scarce scenarios.
 The macro average of F1-score
 is used as the evaluation metric in the CKE-Net model.

The \textbf{BS-RGCN} model proposed by Luo et al. advances stance detection by integrating sentiment and commonsense knowledge through an innovative architecture. It features a graph autoencoder with relational graph convolutional network (RGCN) encoders and a DisMult decoder to capture commonsense knowledge from ConceptNet. Additionally, it incorporates sentiment-aware BERT (SentiBERT) to enhance stance classification. Unlike the CKE-Net model, which achieves state-of-the-art zero-shot stance detection using BERT and ConceptNet but is limited to two-hop knowledge relations, BS-RGCN addresses this limitation by incorporating sentiment and utilizing a broader scope of commonsense knowledge. This comprehensive approach allows BS-RGCN to generalize more effectively across various types of related knowledge, improving performance in both zero-shot and few-shot scenarios. The model’s macro F1 scores of 72.6\% and 71.3\% on the VAST dataset highlight its superior capabilities. BS-RGCN shows enhanced accuracy for stance classification, particularly with sentiment and stance pairs such as (Pos, Pro) and (Neg, Con). Its increased coverage of commonsense knowledge also correlates with better performance, demonstrating its direct benefit for stance detection. Case studies reveal that BS-RGCN effectively handles nuanced cases with implicit topic references and complex sentiment cues, proving its robustness in integrating sentiment and commonsense knowledge. This approach significantly advances the capabilities of zero-shot and few-shot stance detection compared to CKE-Net and other models \citep{luo2022exploiting}.

In their 2022 paper, He et al. introduced \textbf{WS-BERT} (Wikipedia Stance Detection BERT), a model designed to improve stance detection by incorporating background knowledge from Wikipedia. Traditional models often fail to incorporate the necessary background knowledge, resulting in suboptimal performance. WS-BERT addresses this by infusing Wikipedia knowledge into the stance encoding process. The model is evaluated on three benchmark datasets covering social media discussions and online debates, demonstrating significant performance improvements over state-of-the-art methods in target-specific stance detection, cross-target stance detection, and zero/few-shot stance detection \citep{he2022infusing}. WS-BERT leverages different BERT-based models depending on the nature of the text. For formal documents, WS-BERT-Single uses a single BERT model pretrained on Wikipedia to collectively encode the document, target, and Wikipedia knowledge. For informal, noisy texts from social media, WS-BERT-Dual employs two separate language models: BERTweet or COVID-Twitter-BERT for the document-target pair and vanilla BERT for Wikipedia knowledge. This dual encoding minimizes domain shifts between training examples and the pretraining corpora of the language models. Extensive experiments on P-Stance, COVID-19-Stance, and VAST datasets reveal that WS-BERT significantly enhances stance detection accuracy, especially in scenarios where background knowledge is crucial for understanding the target, such as cross-target and zero/few-shot stance detection tasks.

The \textbf{TarBK-BERT} proposed method for Zero-Shot Stance Detection (ZSSD) integrates targeted background knowledge extracted from Wikipedia to enhance the model's ability to generalize stance detection across unseen targets \citep{zhu2022enhancing}. By crawling and filtering relevant knowledge using keyword matching, the approach enriches BERT-based models with target-specific information, facilitating more accurate stance predictions for new targets not seen during training. Specifically, the method leverages pre-trained BERT embeddings fed with context enriched by targeted background knowledge, referred to as TarBK-BERT. This novel approach addresses the ZSSD task by bridging the gap between known and unknown targets through enriched representations, thereby improving overall model performance. For evaluating cross-target stance detection, experiments were conducted on the Sem16 dataset, featuring predefined targets such as Donald Trump, Hillary Clinton, and others. Results demonstrate that TarBK-BERT consistently outperforms baseline models in cross-target scenarios. Specifically, when tested on related but unseen targets, TarBK-BERT achieves significant improvements in stance detection accuracy compared to other state-of-the-art methods. This highlights the effectiveness of integrating targeted background knowledge in enhancing the model's ability to generalize across different targets, underscoring its applicability and robustness in real-world natural language processing tasks.

The \textbf{NPS4SD} (Neural Production System for Stance Detection) is an interpretable end-to-end deep learning model designed to classify the stance of opinionated text towards a specified target. It consists of two main components: a pretrained model for learning text representations and a Variable Binding Network (VBN) that integrates knowledge rules with text entities. The VBN dynamically selects rule-entity patterns, allowing the model to apply relevant rules based on the context.
NPS4SD enhances stance detection through three types of knowledge rules: (1) Target-specific knowledge, which captures information relevant to the given target, (2) Background knowledge from external sources like Wikipedia to provide additional context, and (3) Syntax-related knowledge, utilizing dependency parsing to improve understanding of the text’s structure. By combining these rules with deep learning, NPS4SD addresses key limitations in stance detection, particularly in terms of interpretability and the integration of human-like reasoning into model predictions \citep{zhang2023twitter}. The model was evaluated on three real-world datasets: SemEval-2016, P-stance, and VAST, under various setups including in-domain, cross-target, and zero-shot scenarios. Using the F1-score, a standard metric for classification, results indicate that NPS4SD outperforms state-of-the-art baseline methods, especially in cross-target and zero-shot settings. For cross-target stance detection, NPS4SD improved the F1 score by an average of 9.0\% over methods like TPDG and by 8.6\% compared to RoBERTa, demonstrating its robustness with unseen targets. In zero-shot detection, leveraging background knowledge rules was crucial, leading to significant performance gains. NPS4SD not only improves stance detection accuracy but also enhances model interpretability, offering a more transparent approach compared to traditional deep neural networks.

The \textbf{ANEK} (adversarial network with external knowledge) model addresses the challenge of zero-shot stance detection by incorporating adversarial learning, sentiment information, and common sense knowledge. The model leverages pre-trained models like BERT and SentiBERT to create robust contextual and sentiment representations. A key feature is the use of a graph autoencoder trained on subgraphs from ConceptNet to integrate common sense knowledge into the stance detection process. ANEK also utilizes contrastive learning to enhance the quality of these representations, enabling better generalization to unseen targets. This approach significantly improves upon existing methods by addressing limitations like the transfer of target-invariant information and understanding implicit viewpoints in text \citep{chunling2023adversarial}. ANEK was tested on three datasets: SEM16, WT-WT, and COVID-19. In the cross-target stance detection experiments on the SEM16 dataset, ANEK demonstrated superior performance, particularly when compared to models like BERT and TOAD. The cross-target setting proved more effective than the standard zero-shot setting, suggesting that pre-existing knowledge of relationships between targets helps the model learn more reliable target-invariant representations. This highlights the challenges of zero-shot stance detection and the effectiveness of incorporating external knowledge to enhance the model's understanding and generalization abilities.

\textbf{CNet-Ad} leverages a combination of BERT, a Relational Graph Convolution Network (R-GCN), and a feature separation adversarial network to tackle zero-shot stance detection (ZSSD). BERT encodes text and target embeddings, while the R-GCN captures commonsense knowledge from ConceptNet to address the lack of target context. The feature separation adversarial network distinguishes between target-specific and target-invariant features, enhancing the model's ability to generalize to unseen targets and maintain performance. By combining commonsense knowledge with adversarial learning, CNet-Ad improves performance by ensuring that both types of features—target-specific and target-invariant—are effectively utilized. This approach addresses the limitations of existing methods that either fail to adequately handle unseen targets or rely heavily on noisy external knowledge \citep{zhang2024commonsense}. In the cross-target setting, CNet-Ad was tested on the SemEval-2016 dataset, where it demonstrated significant improvements over baseline methods such as  JointCL and SEKT. Also, CNet-Ad generally outperforms the MPCL model, except in two target pairs (LA→FM and HC→DT) where MPCL achieves better results among all the models. Despite the challenges of handling only two types of targets, CNet-Ad achieved superior results by effectively leveraging both target-specific and target-invariant features. This performance underscores the model's robustness and ability to extend its capabilities to new, unseen targets in cross-target scenarios.

\subsection{Knowledge-enhanced Prompt-tuning based methods:}
\textbf{KEprompt} is a stance detection model that uses a prompt-tuning framework combined with background knowledge to enhance performance. The model architecture leverages pretrained language models such as BERT-base, BERT-large, RoBERTa-base, and RoBERTa-large. Its innovation lies in the automatic verbalizer method, which reduces bias by selecting appropriate label words, and the integration of external knowledge from the SecticNet lexicon and ConceptGraph. The authors also proposed a new dataset named \textbf{ISD} for stance detection tasks. Experiments were conducted on the SEM16, P-stance, ISD, and VAST datasets, using micro-average and macro-average F1-scores for evaluation \citep{huang2023knowledge}. Results showed that KEprompt significantly outperforms baseline methods in both in-domain and cross-target setups. In cross-target stance detection, KEprompt improves F1avg by 8.4\% and F1m by 8.0\% compared to the statistical method TPDG, and surpasses fine-tuning methods by 14.6\% to 11.4\% on average. In zero-shot stance detection, KEprompt demonstrates its effectiveness, achieving superior performance on the VAST dataset compared to all baselines. Ablation studies confirm that the prompt-tuning framework and automatic verbalizer are crucial for the model's success, with the integration of SenticNet enhancing performance through better semantic coverage. These findings underscore KEprompt's robustness and effectiveness across various stance detection tasks, highlighting its significant contribution to the field.

\textbf{INJECT} model proposed by Beck et al. employs a dual-encoder architecture to enhance stance detection by integrating contextual information. The model uses BERT as the foundational language model for encoding both the input text and the additional context. BERT processes the text and context separately, facilitating their interaction through cross-attention mechanisms. Additionally, INJECT leverages T0pp to generate relevant context via prompts, enhancing its ability to handle various stance detection scenarios. It also incorporates ConceptNet and CauseNet for sourcing contextual knowledge, with ConceptNet providing commonsense relationships and CauseNet offering causal connections relevant to stance targets \citep{beck2023robust}. In the cross-target setting, INJECT demonstrates improved performance by generalizing better to unseen targets compared to existing methods. The model's robustness stems from its ability to integrate context flexibly, using multiple sources like ConceptNet, CauseNet, and T0pp, which is beneficial for diverse datasets and noisy contexts. This approach addresses limitations of traditional stance detection models, which often rely too heavily on target-specific vocabulary or struggle with cross-target generalization. Overall, INJECT shows superior performance across a diverse range of datasets, outperforming many state-of-the-art models by effectively leveraging contextual information.

Zhao Zhang et al. introduced \textbf{LKI-BART}, a novel approach for stance detection that excels in zero-shot and cross-target scenarios by leveraging LLM-driven knowledge. The method integrates LLM-driven contextual knowledge into a BART-based generation framework, enhancing its ability to interpret and predict stances towards unseen targets. Additionally, a prototypical contrastive loss is used to align stance representations with semantic labels, improving accuracy and robustness. Unlike traditional methods that focus mainly on target-related background knowledge, LKI-BART incorporates LLM-driven knowledge to explicitly capture the relationship between text and target, enhancing contextual understanding.\citep{zhang2024llm}. LKI-BART was evaluated on the VAST and P-Stance datasets, showing significant performance improvements over existing methods. In cross-target stance detection, LKI-BART demonstrated superior results, with up to 15 F1 points higher than previous models. The approach's effectiveness in both zero-shot and cross-target settings underscores its advanced capability in handling complex stance detection tasks.

In the paper by Bowen Zhang et al., the proposed \textbf{KAI} (knowledge-augmented interpretable network) model integrates two main components: LLM-KE and Bi-KGNPS. The LLM-KE component uses chain-of-thought (CoT) prompting with large language models (LLMs) to generate target-relevant analytical perspectives and rationales. Specifically, it first asks the LLM to enumerate distinct perspectives on the target and then generates rationales for these perspectives. This process involves prompt-based elicitation of knowledge, enabling the model to utilize target-independent transferable knowledge. The Bi-KGNPS component consists of two branches. The left branch processes the input text and target, using perspective features to enhance text representations through a multihop attention mechanism. The right branch processes the perspectives and rationales to construct variable bindings for selecting relevant perspectives and learning content-oriented features. This dual-branch setup facilitates dynamic interaction and knowledge-variable binding, refining the prediction process \citep{zhang2024knowledge}. For evaluation, the KAI model uses accuracy and F1-score average ($F_{\text{avg}}$) as metrics. It achieves state-of-the-art performance in cross-target stance detection on SEM16 dataset, demonstrating significant improvements over baseline models. The model performs well in both zero-shot and few-shot settings, benefiting from its ability to leverage target-independent knowledge and domain-specific insights effectively.

The \textbf{PSDCOT} model addresses significant challenges in stance detection through its innovative use of prompt-based methods combined with advanced knowledge extraction techniques. The model integrates two main components: knowledge extraction via ChatGPT, which employs a chain-of-thought (COT) approach to generate detailed background knowledge, and knowledge fusion through a multi-prompt learning network (M-PLN) using RoBERTa. In the prompt-based method, specific prompts are designed to elicit relevant knowledge from the language model, such as “The attitude towards <Target> is [MASK],” which allows the model to infer stance by filling in the mask with appropriate stance labels. This framework helps mitigate the limitations of traditional fine-tuning approaches, which often struggle with the gap between pre-training and task-specific requirements \citep{ding2024leveraging}. For cross-target stance detection, PSDCOT significantly outperforms other methods, including TPDG (a statistical method) and RoBERTa-FT (a fine-tuning based method). Evaluations on datasets like SemEval-2016 and P-Stance using the micro-average F1 score demonstrate an average improvement of about 16.15\% over statistics-based models like TPDG and 9.73\% over fine-tuning based methods like RoBERTa-FT. In zero-shot stance detection, where the target may be completely new, PSDCOT also excels, particularly on the VAST dataset, by leveraging its prompt-tuning framework and ChatGPT-enhanced knowledge. Despite the inherent challenges of zero-shot settings, PSDCOT effectively incorporates background knowledge and adapts to unseen targets, demonstrating its robustness and versatility.

\section{Summary of Surveyed Studies}
\begin{figure}[h!]
    \centering
    \includegraphics[width=0.5\textwidth]{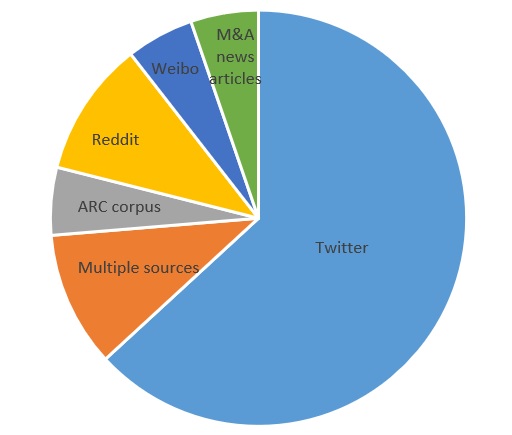}
    \caption{Distribution of platforms used across various stance datasets.}
    \label{figs:pie-chart.jpg}
\end{figure}

\begin{figure}[htb]
	\centering \includegraphics[width=0.6\textwidth]{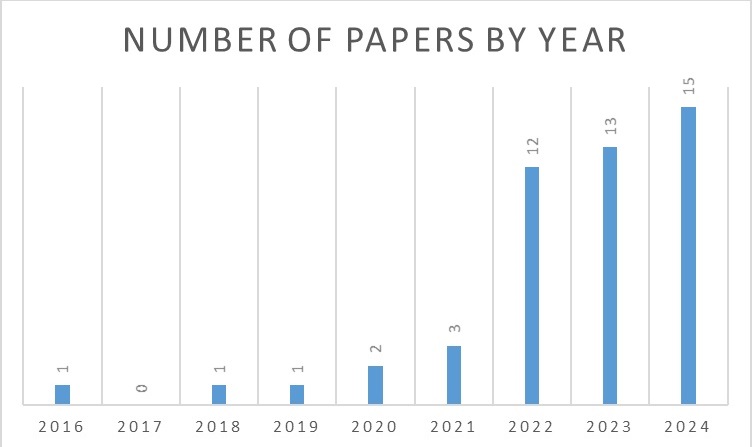}
	\caption{Publications per year up to August 2024.}
	\label{figs:papersnumbers}
\end{figure}

In the end, we gathered 48 relevant papers, as depicted in Figure \ref{figs:papersnumbers}, which illustrates an upward trend in publications from the first in 2016 through 2024. Also, Figure \ref{figs:pie-chart.jpg} illustrates the distribution of platforms utilized in the stance datasets reviewed in section \ref{datasets}. Twitter is the predominant platform, appearing in the majority of datasets, while platforms such as Reddit, Weibo, and ARC corpus are used less frequently.

\section{Open Challenges and Future Directions}

\subsection{Open Challenges}

While there has been significant progress in cross-target stance detection over the last decade, there are still numerous limitations and weaknesses that need further exploration. One major issue is the challenge of generalizing to unseen targets; which is however a frequent, real-world challenge in various domains, such as for example in politics where new political candidates emerge for which data for training a stance detection model is not available. Models such as TPDG and RoBERTa-FT demonstrate this limitation when facing new targets, as they rely heavily on target-specific features and labeled data. These models often underperform in cross-target scenarios due to their dependency on training data that may not encompass the range of target types that will be seen in the future. For instance, in evaluations where models like ANEK showed robust performance, this success often depended on their ability to leverage external knowledge effectively. However, even advanced methods like ANEK, which utilize adversarial learning and sentiment information, sometimes fall short in scenarios where the target is completely new or out-of-distribution. This highlights a significant weakness in current approaches: the reliance on target-specific data or pre-existing knowledge that may not always align with novel or unseen targets.

Another limitation is related to the integration and application of external knowledge. While models like KEprompt and PSDCOT attempt to address this by incorporating various knowledge sources, they still encounter difficulties in dynamically selecting and utilizing the most relevant knowledge for each specific target. KEprompt, for instance, integrates external knowledge from sources like SenticNet and ConceptGraph but may struggle with ensuring that this knowledge is always applicable and relevant across different contexts. Similarly, PSDCOT’s reliance on prompt-tuning and background knowledge, while improving performance, still faces challenges in adapting this knowledge dynamically in zero-shot scenarios. The issue is exacerbated by the complexity of effectively managing and integrating diverse knowledge sources, which can lead to an over-reliance on certain types of information and potentially limit the model’s flexibility and accuracy.

Additionally, the interpretability and transparency of stance detection models remain a concern. Although models like NPS4SD offer improvements in interpretability by integrating knowledge rules, there is still a need for more refined approaches that balance between transparency and performance. The ability to explain why a model made a certain prediction, especially in complex cross-target scenarios, is crucial for trust and understanding. Current models, despite their advancements, often provide limited insight into their decision-making processes, making it challenging to diagnose and address their weaknesses effectively.

\subsection{Avenues for Future Work}

To advance cross-target stance detection, future research should focus on enhancing generalization capabilities by developing models that can better adapt to novel and unseen targets. This involves improving techniques for leveraging both target-specific and target-invariant features in a more balanced and dynamic manner. One promising avenue is the exploration of more sophisticated knowledge integration methods, such as adaptive knowledge retrieval systems that can dynamically adjust based on the target's context. Incorporating mechanisms for real-time knowledge updates could also help models stay current with emerging targets and contexts, enhancing their robustness and relevance.

Additionally, enhancing model interpretability should be a priority. Future work could focus on developing methods that provide clearer insights into how models make decisions, particularly in complex cross-target scenarios. Techniques such as explainable AI (XAI) and transparent reasoning frameworks could be integrated to offer more detailed explanations of model predictions. This would not only improve user trust and understanding but also facilitate more effective model debugging and refinement. Combining these advances with improved knowledge integration strategies could significantly enhance the performance and applicability of stance detection models across diverse and evolving scenarios.

Where large language models (LLMs) are increasingly being used in natural language processing \citep{zubiaga2024natural}, their use for stance detection is still in its infancy \citep{lan2024stance}. However, the limited research to date has shown that they do provide a promising avenue for research towards generalization in stance detection \citep{wagner2024power,mahmoudi2024zero}, which remains largely unexplored in cross-target stance detection \citep{ding2024cross,zhang2024llm}.

\section{Conclusion}

This survey provides a comprehensive overview of recent advancements and challenges in cross-target stance detection, highlighting the evolution of models and methodologies as well as existing datasets suitable for the task. It discusses 48 papers introducing methods to cross-target stance detection, as well as 15 datasets. It reviews key approaches such as NPS4SD, ANEK, CNet-Ad, and knowledge-enhanced prompt-tuning methods like KEprompt, INJECT, LKI-BART, KAI, and PSDCOT. Each of these methods brings unique contributions to the field, addressing different aspects of stance detection, from integrating external knowledge and commonsense reasoning to improving zero-shot and cross-target performance.

To conclude, the review critically analyzes the current state of the field, delving into the key open challenges and the main avenues suggested for future research.

\printcredits

\section*{Acknowledgments}
Parisa Jamadi Khiabani is funded by the Islamic Development Bank (IsDB).\\ 

Declaration of generative AI and AI-assisted technologies in the writing process:
During the preparation of this work, the author(s) used ChatGPT in order to improve writing clarity. After using this tool/service, the author(s) reviewed and edited the content as needed and take(s) full responsibility for the content of the publication.
\bibliographystyle{cas-model2-names}

\bibliography{cas-refs}





\end{document}